\documentclass[a4paper,fleqn, review]{cas-dc}
\usepackage{amsmath}
\usepackage[switch]{lineno}
\usepackage{graphicx}
\usepackage{subfig} 
\usepackage{tabu}  
\usepackage{hyperref}
\usepackage{multirow}
\usepackage{dsfont}
\usepackage{stfloats}
\usepackage{cite}
\usepackage{booktabs}
\usepackage{algorithm}
\usepackage{algorithmic}
\usepackage{amssymb}
\usepackage{makecell}
\usepackage{caption}
\usepackage{adjustbox}
\usepackage{enumitem}
\usepackage{natbib}
\usepackage{booktabs}
\usepackage{siunitx}

\def\tsc#1{\csdef{#1}{\textsc{\lowercase{#1}}\xspace}}
\tsc{WGM}
\tsc{QE}
\tsc{EP}
\tsc{PMS}
\tsc{BEC}
\tsc{DE}

\begin{document}
\let\WriteBookmarks\relax
\def\floatpagepagefraction{1}
\def\textpagefraction{.001}
\shorttitle{}

\shortauthors{Jian Song et~al.}

\title [mode = title]{Enhancing Monocular Height Estimation via Sparse LiDAR-Guided Correction}                      



%
\author[1,2]{Jian Song}[orcid=0009-0001-5577-8595]

\author[1,2]{Hongruixuan Chen}





\author[1,2]{Naoto Yokoya}
\cormark[1]

\affiliation[1]{organization={Graduate School of Frontier Sciences, The University of Tokyo},
    city={Chiba},
    postcode={277-8561}, 
    country={Japan}}



\affiliation[2]{organization={RIKEN Center for Advanced Intelligence Project (AIP), RIKEN},
    city={Tokyo},
    postcode={103-0027}, 
    country={Japan}}



\cortext[cor1]{Corresponding author}


\nonumnote{Manuscript submitted on November 7, 2025.}

\begin{abstract}
Monocular height estimation (MHE) from very-high-resolution (VHR) remote sensing imagery via deep learning is notoriously challenging due to the lack of sufficient structural information. Conventional digital elevation models (DEMs), typically derived from airborne LiDAR or multi-view stereo, remain costly and geographically limited. While state-of-the-art monocular height estimation (MHE) and depth estimation (MDE) models show great promise, their robustness under varied illumination conditions remains a significant challenge. To address this, we introduce a novel and fully automated correction pipeline that integrates sparse, imperfect global LiDAR measurements (ICESat-2) with deep learning outputs to enhance local accuracy and robustness. Importantly, the entire workflow is fully automated and built solely on publicly available models and datasets, requiring only a single georeferenced optical image to generate corrected height maps, thereby ensuring unprecedented accessibility and global scalability. Furthermore, we establish the first comprehensive benchmark for this task, evaluating a suite of correction methods that includes two random forest-based approaches, four parameter-efficient fine-tuning techniques, and full fine-tuning. We conduct extensive experiments across six large-scale, diverse regions at 0.5\,m resolution, totaling approximately 297\,km², encompassing the urban cores of Tokyo, Paris, and São Paulo, as well as mixed suburban and forest landscapes. Experimental results demonstrate that the best-performing correction method reduces the MHE model's mean absolute error (MAE) by an average of 30.9\% and improves its $F_1^{\text{HE}}$ score by 44.2\%. For the MDE model, the MAE is improved by 24.1\% and the $F_1^{\text{HE}}$ score by 25.1\%. These findings validate the effectiveness of our correction pipeline, demonstrating how sparse real-world LiDAR data can systematically bolster the robustness of both MHE and MDE models and paving the way for scalable, low-cost, and globally applicable 3D mapping solutions.
\end{abstract}

\begin{keywords}
\sep Monocular Height Estimation \sep 
ICESat-2 Data \sep Sparse LiDAR-based Calibration \sep Parameter-efficient fine-tuning  \sep Machine Learning

\end{keywords}

\maketitle

\section{Introduction}

Very-high-resolution (VHR) sensors enable increasingly detailed observations of the Earth, providing diverse data modalities that enrich our understanding of surface conditions. For instance, the WorldView\footnote{\url{https://earth.esa.int/eogateway/missions/worldview-3}} satellite series can now offer sub-meter ground-sampling distance (GSD) optical imagery, as well as multispectral/hyperspectral data\footnote{\url{https://www.ehu.eus/ccwintco/index.php?title=Hyperspectral_Remote_Sensing_Scenes}} (capturing multiple spectral bands for enhanced material and vegetation analysis) and synthetic aperture radar (SAR) data (capable of all-weather, day-and-night imaging)~\citep{chen2025bright,xia2025openearthmap}. Among these modalities, VHR digital elevation models (DEMs), which capture the elevation of terrain and above-ground objects, play a pivotal role in urban planning, environmental monitoring, disaster management, 3D mapping, and digital twin applications~\citep{li20193d, 3dbuilidng1, 3dbuilding2, 3dbuilding3, building_height2,building_height}.

Although moderate- to low-resolution DEMs (30 m or coarser), such as SRTM~\citep{SRTM}, ASTER~\citep{ASTER}, ALOS PALSAR~\citep{ALOS_PALSAR}, and TanDEM~\citep{TanDEM_X}, are freely available on a global scale, obtaining sub-meter DEMs traditionally relies on methods like airborne LiDAR~\citep{sohn2004extraction,hermosilla2011evaluation,li2020high}, stereo vision matching~\citep{ameri2002high, zhang2003multi, multiview3d_1, liu2023deep, han2020state, mahphood2019dense}, or InSAR~\citep{wang2024domain,rs71215809}—techniques that are both expensive and time-consuming. For example, based on AW3D’s\footnote{\url{https://net.jmc.or.jp/mapdata/3d/aw3d/enhanced.html}} per-square-kilometer pricing, generating a 0.5\,m GSD DEM of Japan through stereo matching could cost up to 20 million U.S. dollars. Similarly, the French HD LiDAR project\footnote{\url{https://diffusion-lidarhd.ign.fr/mnx/}} estimates that acquiring nationwide LiDAR coverage of France would require an investment of nearly 60 million euros and take about five years. Such high costs and limited scalability have hindered the widespread global adoption of high-resolution DEM applications.

The recent rise of deep learning offers a promising alternative. Researchers have explored using machine learning, particularly \emph{monocular height estimation} (MHE), to infer elevations from a single VHR remote sensing optical image, greatly reducing costs and allowing for broad scalability ~\citep{joint_height, joint_height4, jonint_height2, joint_height3,li20193d, 3dbuilidng1, 3dbuilding2, 3dbuilding3, building_height2, gordon2020learning, img2dsm, building_height}. Yet, like traditional approaches, deep learning models require extensive labeled data, which remains difficult to obtain at sub-meter resolution on a global scale. Moreover, remote sensing data often exhibit geographic biases—abundant in developed regions but scarce elsewhere—leading to inductive bias in models trained predominantly on real-world data~\citep{schmitt2023there}, thereby limiting the overall applicability of MHE to different regions.

Alongside MHE, \emph{monocular depth estimation} (MDE) from the computer vision field has recently been explored for height estimation tasks~\citep{cambrin2024depth}. However, MDE models present critical limitations for this application. First, they are typically trained on natural, ground-level images where objects exhibit rich structural information, a feature largely absent in overhead-view remote sensing imagery. Second, they output relative depth rather than absolute metric heights, making their predictions ambiguous without a reliable external anchor to provide a correct scale and offset.

Despite the potential of MHE and MDE models, their reliability under diverse real-world conditions remains a critical concern. For human observers, inferring height from a single overhead image heavily relies on cues like shadows. Whether neural networks adopt a similar, potentially fragile, mechanism is poorly understood. This lack of understanding is risky, as factors like sun angle and weather can dramatically alter an image's appearance, possibly leading to significant prediction errors. Moreover, unlike tasks such as semantic segmentation, MHE outputs cannot be easily validated by human inspection, making it difficult to spot subtle but critical elevation errors.

To systematically investigate these potential vulnerabilities, we designed a controlled experiment by building a synthetic environment \citep{song2024synrs3d}. By simulating the same scene under varied illumination and texture conditions, we could isolate the impact of these factors on model predictions. We tested a state-of-the-art MHE model, RS3DAda~\citep{song2024synrs3d}, and a leading MDE model, Depth Anything V2~\citep{yang2024depthv2}. Our findings reveal a critical flaw: both models are highly sensitive to shadow variations, producing inconsistent and systematically biased height estimations as illumination changes. This discovery confirms that while these models can generate structurally plausible dense outputs, their absolute accuracy is fundamentally unreliable.

This identified vulnerability directly motivates the need for a post-processing correction step. While generating a dense, sub-meter ground truth DEM for correction is prohibitively expensive (the very problem we aim to solve), globally available sparse elevation data offers a highly practical alternative. Instruments like NASA's GEDI\footnote{\url{https://gedi.umd.edu/}} and ICESat-2\footnote{\url{https://icesat-2.gsfc.nasa.gov/}} provide high-accuracy, albeit sparse, height measurements. These sparse points can act as an anchor to correct the systematic biases of the dense but unreliable height maps generated by deep learning models.

Building on this insight, we propose an automated post-processing correction pipeline that leverages ICESat-2 data to refine dense height predictions. This workflow comprises two main steps: robust preprocessing of raw ICESat-2 data, followed by a correction stage. In this stage, we benchmark a wide array of methods, including not only traditional machine learning but also modern \emph{parameter-efficient fine-tuning} (PEFT) techniques, which adapt large pre-trained models with minimal computational cost.

Our key contributions are as follows:
\begin{enumerate}
    \item We design and validate a novel, fully automated post-processing pipeline that leverages sparse ICESat-2 data to significantly improve the accuracy of height maps generated by both state-of-the-art MHE and MDE models.
    \item We establish the first comprehensive benchmark of correction methods for this task, systematically evaluating traditional machine learning, multiple parameter-efficient fine-tuning techniques, and full fine-tuning. Such kind of pipeline and benchmark will facilitate the research in the relevant communities.
    \item We conduct an extensive, large-scale evaluation across approximately 297\,km² of diverse urban and rural landscapes, demonstrating the robustness and generalizability of our proposed pipeline.
    \item We highlight the unprecedented accessibility and scalability of our pipeline: it is fully automated, relies exclusively on open and globally available resources (e.g., ICESat-2, FABDEM, and open-source models), and requires only a single georeferenced optical image to operate, enabling truly global, low-cost 3D mapping.
\end{enumerate}

The remainder of this paper is organized as follows. Section~\ref{sec:relatedwork} reviews related work. Section~\ref{sec:height_mechanism} provides the shadow-based analysis that  directly motivates this work. Section~\ref{sec:studyareas} introduces the study areas and the ICESat-2 data. Section~\ref{sec:icesat2_calibration} details our core contribution: the end-to-end correction pipeline, including data preprocessing and the benchmark of calibration strategies. Section~\ref{sec:benchexperiments} presents the experimental results, followed by a discussion and conclusion in Section~\ref{sec:limitations_discussion}.

\section{Related Work}
\label{sec:relatedwork}

This study spans several interconnected research domains, including monocular depth estimation (MDE) in computer vision, monocular height estimation (MHE) in remote sensing, parameter-efficient fine-tuning (PEFT) for adapting foundation models, and the utilization of ICESat-2 data for elevation mapping. Our overarching aim is to achieve accurate and reliable high-resolution height mapping without relying on costly, dense real-world datasets such as airborne LiDAR or photogrammetry. In the following subsections, we systematically review related work across these four areas.

\subsection{Monocular Depth Estimation}
Monocular depth estimation (MDE) has been extensively studied in the computer vision community, primarily for natural images. Early approaches relied on handcrafted features and probabilistic graphical models~\citep{saxena2005learning}, but the advent of deep learning has led to substantial progress. \cite{eigen2014depth} first demonstrated the feasibility of predicting dense depth maps from a single RGB image using multi-scale CNNs. Subsequent works introduced encoder-decoder architectures, residual learning, and conditional random fields to further enhance accuracy~\citep{laina2016deeper,liu2015learning}. More recent advances leverage large-scale datasets and transformer-based backbones, improving the transferability of representations~\citep{zhao2022monovit,ranftl2021vision}.

State-of-the-art MDE models are increasingly pretrained on diverse image corpora to boost generalization. For example, the MiDaS framework~\citep{ranftl2020towards} unifies multiple datasets into a single training pipeline, while Depth Anything V2~\citep{yang2024depth} leverages strong self-supervised pretraining on synthetic natural images to produce highly generalizable depth predictions across varied environments. However, a fundamental limitation persists: these models are trained to predict relative depth. Their outputs, while internally consistent, suffer from an inherent scale and shift ambiguity, meaning they lack a true, absolute metric scale. This ambiguity renders them insufficient for most remote sensing applications, where the primary objective is to derive precise, georeferenced measurements. Our work addresses this gap by introducing a correction pipeline that aligns such relative depth predictions with sparse but globally available ICESat-2 measurements, yielding reliable absolute heights in VHR imagery.

\subsection{Monocular Height Estimation}
Compared with traditional elevation data acquisition methods, such as airborne LiDAR~\citep{sohn2004extraction,hermosilla2011evaluation,li2020high}, stereo matching~\citep{ameri2002high,zhang2003multi,multiview3d_1,liu2023deep,han2020state,mahphood2019dense}, or InSAR~\citep{wang2024domain,rs71215809}, deep learning-based monocular height estimation (MHE) is more cost-effective and scalable. Similar to depth estimation in computer vision, MHE can be categorized into multi-view~\citep{multiview3d_5,multiview3d_4,multiview3d_3,multiview3d_2,multiview3d_1,multiview3d_6} and single-view approaches. While multi-view methods leverage multiple images, single-view approaches only require one, greatly reducing data costs but increasing task difficulty.

Given a high-resolution optical image, the objective is to predict a per-pixel above-ground height map. Earlier works explored multi-task learning to jointly estimate height and semantics~\citep{joint_height4}, or applied residual CNNs validated in instance segmentation~\citep{mou2018im2height}. Generative methods, such as cGANs, formulated MHE as image-to-image translation~\citep{img2dsm}, while others incorporated semantic priors~\citep{joint_height3} or focused on large-scale transfer learning~\citep{xiong2023benchmark}. Recent advances leverage Transformers~\citep{vaswani2017attention}, leading to models such as RS3DAda~\citep{song2024synrs3d}, which couples DINOv2~\citep{oquab2023dinov2} with DPT~\citep{ranftl2021vision} for state-of-the-art results. Beyond generic pipelines, fine-grained tasks like LIGHT~\citep{mao2023light} and GABLE~\citep{sun2024gable} highlight the potential of MHE for national-scale 3D building modeling, while receptive field fusion strategies~\citep{mao2022beyond} further enhance the representation of vertical structures. Our proposed post-processing pipeline builds on these foundations, refining RS3DAda predictions with sparse ICESat-2 supervision.

\subsection{Parameter-Efficient Fine-Tuning}
Large-scale vision transformers and foundation models have shown remarkable capability in representing visual information, but their deployment for specialized downstream tasks is often hampered by the prohibitive cost of full fine-tuning. Parameter-efficient fine-tuning (PEFT) has emerged as a practical alternative, aiming to adapt large models to new tasks by updating only a small subset of parameters.

Several representative PEFT strategies have been proposed. BitFit~\citep{zaken2021bitfit} tunes only the bias terms while freezing all other weights, achieving competitive performance with negligible parameter overhead. Visual Prompt Tuning (VPT)~\citep{jia2022visual} introduces learnable tokens into the input sequence of a transformer, effectively steering the model toward new tasks without modifying its backbone. Adapter methods, such as AdapterFormer~\citep{chen2022adaptformer}, insert lightweight bottleneck layers within transformer blocks, enabling task adaptation with modest additional parameters. Finally, Low-Rank Adaptation (LoRA)~\citep{hu2022lora} decomposes weight updates into low-rank matrices, striking a balance between efficiency and expressivity.

These PEFT methods have been successfully applied to vision tasks ranging from classification to dense prediction, showing that carefully constrained adaptation can outperform full fine-tuning in low-data regimes. In this study, we systematically benchmark four representative PEFT methods (BitFit, VPT, Adapter, and LoRA) for the task of ICESat-2-based calibration, providing new insights into their effectiveness in correcting sparse-data-driven monocular height estimation.

\subsection{ICESat-2 Data}
Launched by NASA in 2018, the Ice, Cloud, and Land Elevation Satellite-2 (ICESat-2) employs the Advanced Topographic Laser Altimeter System (ATLAS) to provide high-precision surface elevation measurements globally. Compared with other spaceborne LiDAR missions (e.g., GEDI), ICESat-2 delivers broader coverage, shorter revisit intervals, and denser along-track sampling, making it well-suited for worldwide elevation applications.

Existing studies have leveraged ICESat-2 to estimate building or forest canopy heights~\citep{shendryk2022fusing,qi2016combining,schneider2020towards,dubayah2022gedi,lao2021retrieving,zhao2023combining,wu2023utilizing,huang2024urban}, but most are limited in spatial scope or rely on auxiliary datasets, restricting large-scale deployment. Moreover, ICESat-2 sampling remains sparse, posing challenges in capturing complex vertical structures in urban areas. Recent efforts fuse ICESat-2 with other modalities~\citep{zhang2019tibetan,li2020high,tang2025flexible}, though these often depend on low- or medium-resolution datasets inadequate for sub-meter mapping.

To address these limitations, we propose a post processing correction pipeline that integrates dense predictions from the state-of-the-art MHE and MDE models with sparse ICESat-2 measurements using a random forest~\citep{breiman2001random}. Our method requires only a georeferenced VHR optical image, regional ICESat-2 tracks, and model outputs, enabling globally applicable, high-fidelity nDSM generation.

\section{Motivation: Model Instability and the Need for Correction}
\label{sec:height_mechanism}
This section provides a diagnostic analysis of model behavior to motivate our correction pipeline, rather than representing a technical contribution of the pipeline itself.

Estimating object heights from a single remote sensing image is notably challenging. While humans can easily recognize objects and their categories from appearance, accurately inferring object heights from a single view is far more difficult. Surprisingly, deep learning models have demonstrated strong performance in MHE. This capability raises fundamental questions: which visual cues do these models prioritize, and how robust is their reliance on these cues under diverse real-world conditions? 

To investigate these questions, we conducted a series of visualization experiments on two state-of-the-art models: a leading MHE method proposed in \citep{song2024synrs3d}, hereafter referred to as the ``MHE model'', and a leading MDE model, Depth Anything V2~\citep{yang2024depth}, hereafter the ``MDE model''. Our investigation reveals a critical dependency: both models heavily rely on shadows to infer height. However, this reliance proves to be a double-edged sword, as we demonstrate that their performance degrades significantly when shadow conditions deviate from those seen during training.

\subsection{Analysis of Model Dependency on Shadow Cues}
\begin{figure}[ht!]
    \centering
    \includegraphics[width=\linewidth]{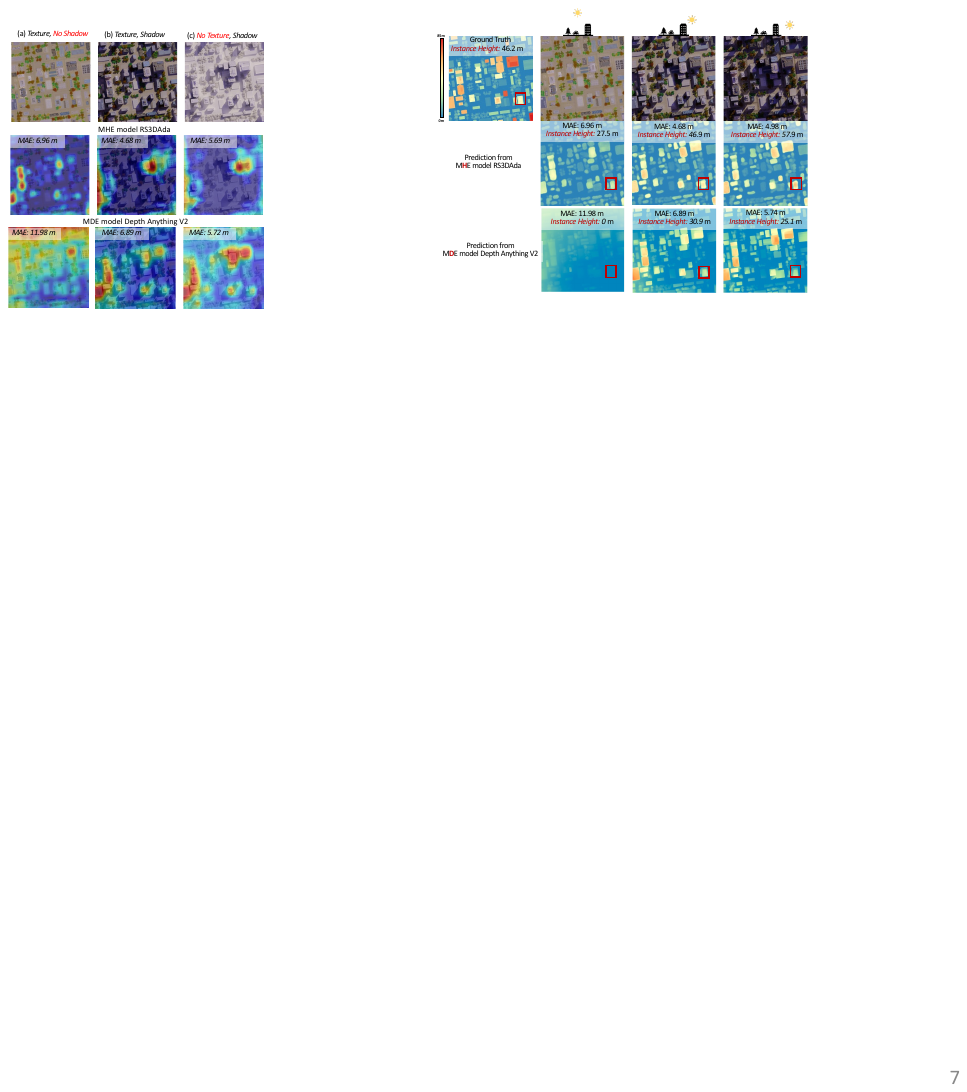}
    \caption{Visualization of three lighting/texture conditions (top row) and the corresponding Grad-CAM attention maps with MAE values for two models (middle: MHE model RS3DAda, bottom: MDE model Depth Anything V2).}
    \label{fig:height_exploration1}
\end{figure}
\par To disentangle the influence of shadows from other visual features, we leveraged a procedural city-synthesis system \citep{song2024syntheworld, song2024synrs3d} to generate three controlled variations of an identical urban scene (Figure~\ref{fig:height_exploration1}, top row): (1) \emph{with texture but no shadows}, (2) \emph{with both texture and shadows}, and (3) \emph{with shadows but no texture}. All three configurations share the same ground-truth height map. Two state-of-the-art models were applied: the RS3DAda model for monocular height estimation (MHE) and the Depth Anything v2 model for monocular depth estimation (MDE). Since MDE produces only relative depth, its predictions were linearly fitted to absolute values using simulated ICESat-2 tracks on the ground-truth data to enable fair evaluation.

\par The results demonstrate a clear and consistent pattern for both models.
\begin{itemize}
    \item Performance is weakest in the shadowless condition (MAE of 6.96 m for MHE, 11.98 m for MDE), where Grad-CAM~\citep{selvaraju2017grad} visualizations show the models focusing primarily on building rooftops.
    \item The introduction of shadows (condition 2) brings a dramatic improvement in accuracy (MAE reduced to 4.68 m and 6.89 m, respectively), and the attention maps shift decisively toward the shadowed areas.
    \item Most revealingly, even when building textures are removed (condition 3), the models maintain strong performance by relying solely on shadow geometry.

\end{itemize}

\par This experiment establishes that shadows serve as the primary and most critical cue for height estimation, far outweighing the influence of surface texture.

\subsection{Quantifying Metric Errors from Illumination Variance}
\begin{figure}[ht!]
    \centering
    \includegraphics[width=\linewidth]{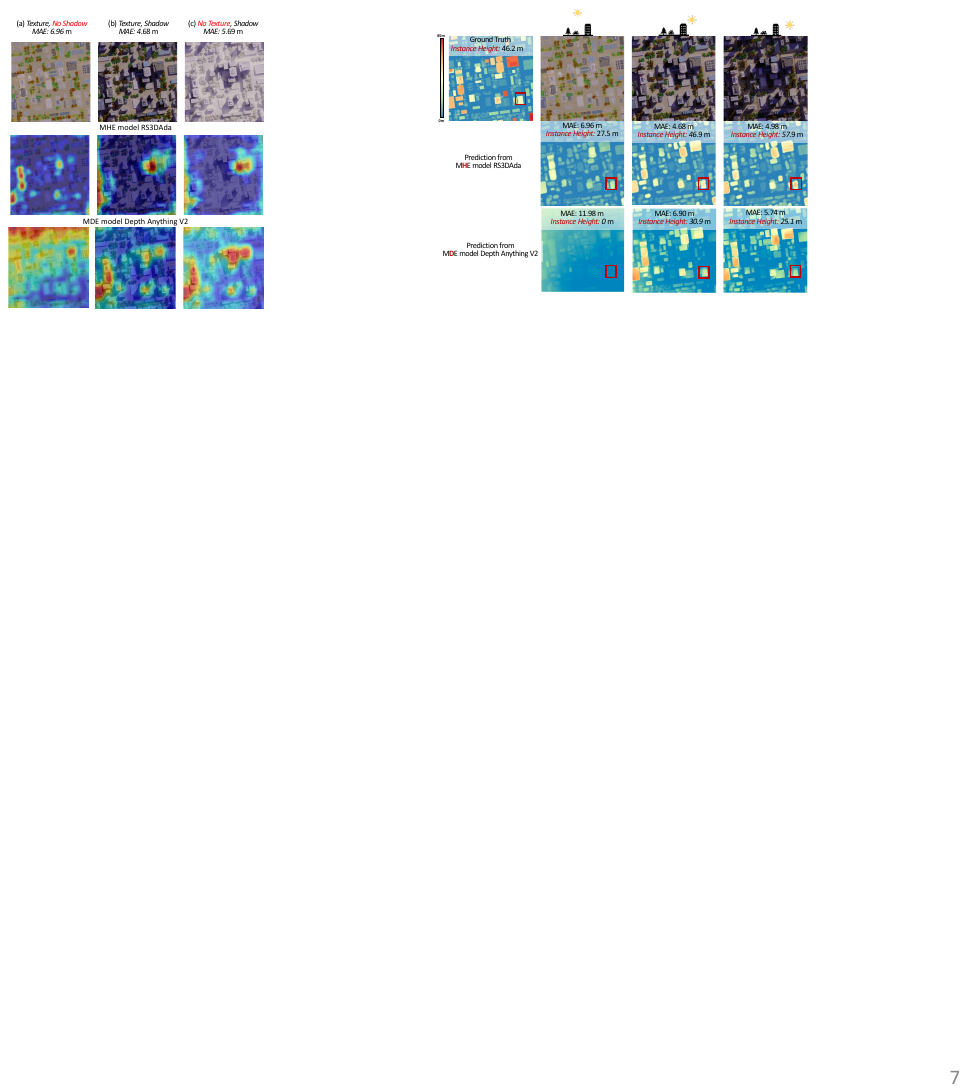}
    \caption{Effect of varying shadow lengths on height estimation.}
    \label{fig:height_exploration2}
\end{figure}
\par Having established that shadows are the dominant cue, we next investigated the models' robustness to variations in them. We simulated three different sun positions to cast (1) \emph{minimal shadows}, (2) \emph{moderate shadows (akin to a typical training condition)}, and (3) \emph{long shadows}, while keeping the scene geometry constant. This experiment effectively serves as a test for the models' generalization capabilities.

All other scene parameters remained fixed, and the ground-truth height map was unchanged. We compared a domain-specific MHE model (RS3DAda) with a large-scale pre-trained MDE model (Depth Anything V2).  

As shown in Figure~\ref{fig:height_exploration2}, both models strongly rely on shadow cues: deviations from training conditions lead to large errors. Each performs best under settings closest to its training regime—the MHE model under moderate shadows, and the MDE model under longer shadows.  

In summary, our analysis leads to a critical insight: while shadows are essential for MHE/MDE models, they are not a stable or generalizable feature. The MHE model's performance is brittle and tied to specific training domains, while the MDE model's priors are ill-suited for overhead imagery. This finding pinpoints shadow-induced error as a fundamental and systematic challenge for monocular height estimation. This vulnerability is precisely what motivates the development of a post-processing correction pipeline, as detailed in the following section, to anchor the models' geometrically plausible but metrically unreliable predictions to a sparse set of accurate ground-truth measurements.

\section{Study Areas and ICESat-2 for Height Calibration}
\label{sec:studyareas}
\begin{figure*}[ht!]
    \centering
    \includegraphics[width=\textwidth]{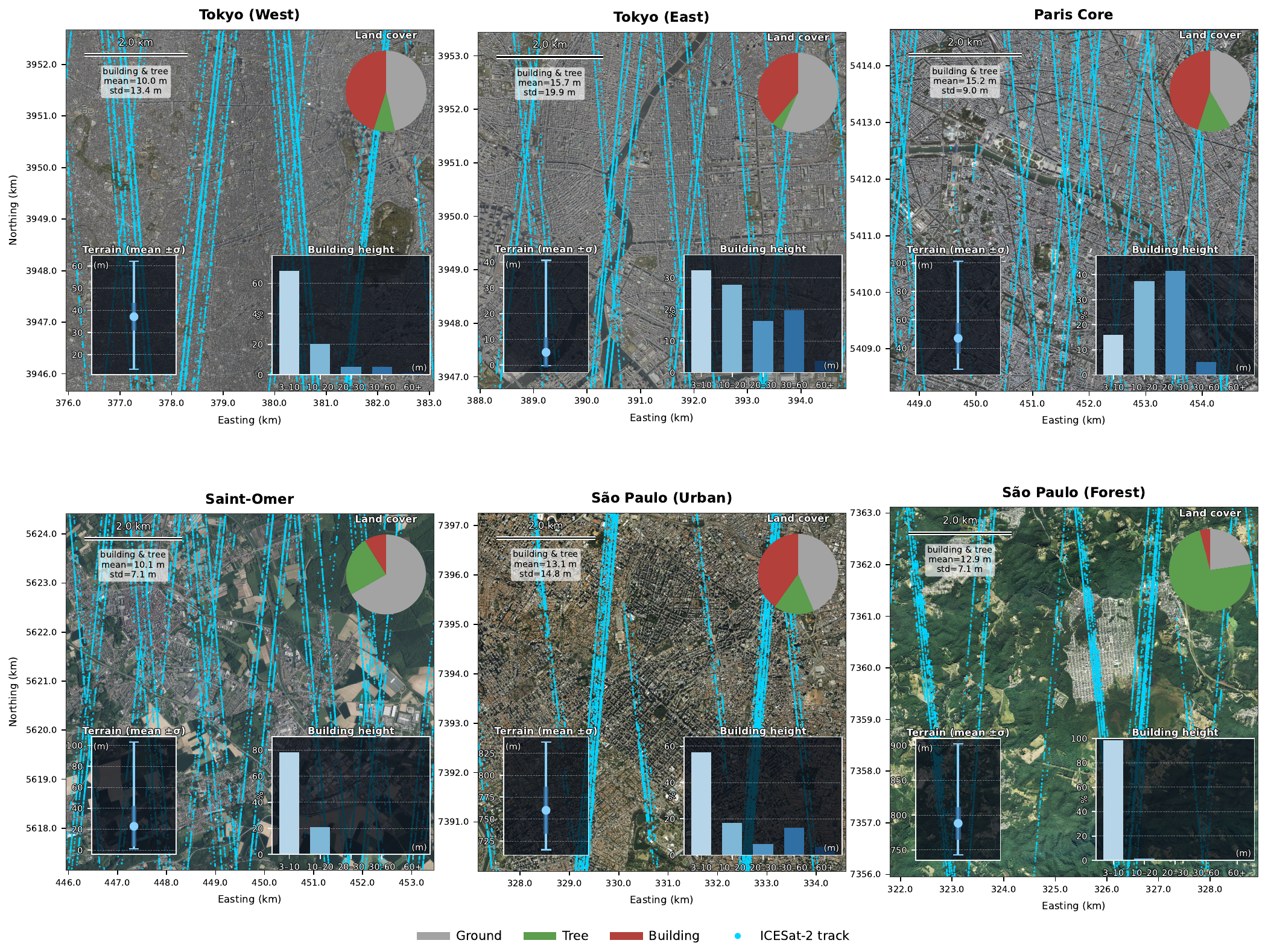}
    \caption{Study areas with overlaid ICESat-2 tracks, illustrating land-cover composition, terrain elevation, building-height distributions, and the mean and standard deviation of above-ground object heights across selected urban, peri-urban, and forested regions.}
    \label{fig:area1}
\end{figure*}

Building on the mechanism analysis in Section~\ref{sec:height_mechanism}, MHE or MDE models perform well on large-scale optical imagery height estimation but often rely excessively on shadow cues, leading to unstable predictions under atypical illumination. Moreover, their \emph{continuous} outputs are inherently more difficult to validate than discrete classification results.

Unlike classification tasks (e.g., ``tree'' vs.\ ``water''), where labels can be visually cross-checked, continuous height predictions cannot be intuitively verified (e.g., distinguishing 3\,m from 7\,m or 16\,m). External reference data are therefore required to expose biases and anchor predictions.  

Sparse but accurate LiDAR missions such as GEDI and ICESat-2 provide such reference signals. Despite their limited coverage, these measurements can reveal systematic errors (e.g., shadow-induced biases) and support global-scale calibration. This section introduces the study areas and outlines the ICESat-2 parameters that enable such correction.

\subsection{Study Areas and Data Sources}
\begin{table*}[ht]
    \centering
    \caption{Key specifications of the six study areas.}
    \label{tab:city_data_specs}
    \adjustbox{max width=0.9\textwidth}{%
    \begin{tabular}{lcccccc}
    \toprule
    \textbf{Region} & \textbf{GSD (m)} & \textbf{Image size (px)} & \textbf{Area (km$^2$)} & \textbf{ICESat-2 overpass} & \textbf{Data source} \\
    \midrule
    \multicolumn{6}{l}{\textit{France}} \\
    Paris Core (France)     & 0.5 & 13003$\times$12776 & 41.53 & 2019.01.01--2024.12.12 & IGN LiDAR HD \\
    Saint-Omer (France)     & 0.5 & 15009$\times$14545 & 54.58 & 2019.01.01--2024.12.12 & IGN LiDAR HD \\
    \addlinespace
    \multicolumn{6}{l}{\textit{Japan}} \\
    Tokyo (East, Japan)     & 0.5 & 13759$\times$13331 & 45.86 & 2019.01.01--2024.12.12 & Tokyo Digital Twin Project \\
    Tokyo (West, Japan)     & 0.5 & 14275$\times$14016 & 50.02 & 2019.01.01--2024.12.12 & Tokyo Digital Twin Project \\
    \addlinespace
    \multicolumn{6}{l}{\textit{Brazil}} \\
    S\~ao Paulo (Urban, Brazil)  & 0.5 & 14904$\times$14507 & 54.05 & 2019.01.01--2024.12.12 & GeoSampa platform \\
    S\~ao Paulo (Forest, Brazil) & 0.5 & 14255$\times$14323 & 51.04 & 2019.01.01--2024.12.12 & GeoSampa platform \\
    \midrule
    \textbf{Total}          & --  & --                 & \textbf{297.08} & -- & -- \\
    \bottomrule
    \end{tabular}
    }
\end{table*}
To evaluate the generalizability of the proposed calibration pipeline, we selected six representative areas across three continents (Europe, Asia, and South America), covering dense metropolitan cores, peri-urban neighborhoods, and sparsely populated mountainous forests (total area: $297.08\,\mathrm{km}^2$). Figure~\ref{fig:area1} visualizes the areas with overlaid ICESat-2 ground tracks, along with summary plots of land-cover composition, terrain elevation, building-height distributions, and the mean and standard deviation of above-ground object heights.

\begin{itemize}
    \item \textbf{Paris Core, France:} 
    The historical center of Paris characterized by dense mid-rise and high-rise structures. Optical imagery at $0.5\,\mathrm{m}$ GSD and reference nDSM are provided by the IGN LiDAR HD project\footnote{\url{https://diffusion-lidarhd.ign.fr/mnx/}}, acquired in 2023.  

    \item \textbf{Saint-Omer, France:} 
    A mid-sized town in northern France featuring moderate building heights interspersed with agricultural and vegetated zones. The reference nDSM is also derived from the IGN LiDAR HD project at $0.5\,\mathrm{m}$ GSD.  

    \item \textbf{Tokyo East and Tokyo West, Japan:}  
    Two adjacent subregions of Tokyo representing the eastern and western metropolitan cores. Both regions include highly heterogeneous building patterns ranging from skyscrapers to dense low-rise blocks. Optical imagery and $0.5\,\mathrm{m}$ GSD nDSMs are sourced from the Tokyo Digital Twin Project\footnote{\url{https://info.tokyo-digitaltwin.metro.tokyo.lg.jp/}}, captured in 2023.  

    \item \textbf{S\~ao Paulo Urban, Brazil:}  
    The dense core of the S\~ao Paulo metropolitan region, one of the largest cities in Latin America, characterized by extensive high-rise clusters and mixed residential-commercial areas. LiDAR data were released by the GeoSampa platform\footnote{\url{https://geosampa.prefeitura.sp.gov.br/PaginasPublicas/_SBC.aspx}} with a point density of $\sim10$ points/m$^2$.  

    \item \textbf{S\~ao Paulo Forest, Brazil:}  
    A peri-urban forested zone located at the boundary of Parelheiros and Marsilac, representing rural and natural terrain with sparse settlement. Reference nDSM is also derived from the GeoSampa LiDAR dataset.  
\end{itemize}

\subsection{ICESat-2 Mission Overview}
\begin{figure}[h!]
\centering
\includegraphics[width=\linewidth]{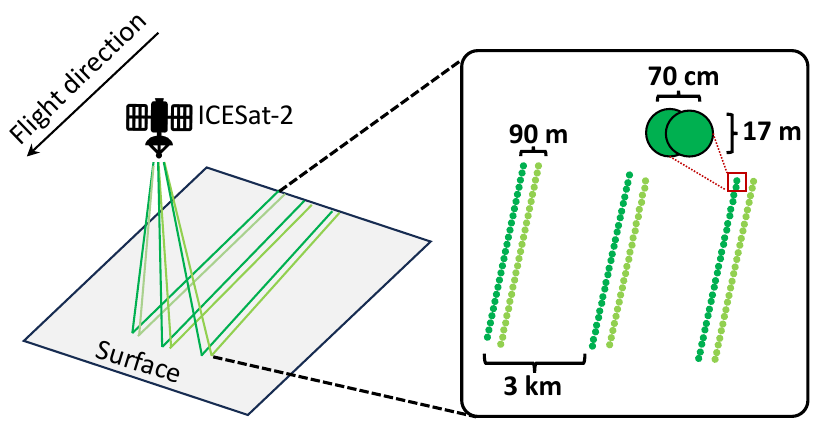}
\caption{ICESat-2 satellite specifications.}
\label{fig:icesat2_specification}
\end{figure}

Figure~\ref{fig:icesat2_specification} summarizes the key technical specifications of the ICESat-2 satellite. 
In addition to these parameters, ICESat-2 provides several advantages for large-scale height calibration:
\begin{enumerate}
    \item \textbf{Near-global coverage:} orbit up to $\pm 88^\circ$ latitude, extending beyond GEDI’s range.  
    \item \textbf{High vertical accuracy:} photon-level ATL03 data achieve sub-meter accuracy under optimal conditions.  
    \item \textbf{Reliable geolocation:} $\pm6.5\,\mathrm{m}$ horizontal accuracy ensures consistent alignment with optical imagery.  
\end{enumerate}

\begin{table}[htbp]
\centering
\caption{Comparison of key parameters for GEDI and ICESat-2.}
\label{tab:gedi_icesat2_specs}
\resizebox{\columnwidth}{!}{%
\begin{tabular}{lll}
\toprule
\textbf{Parameter} & \textbf{GEDI} & \textbf{ICESat-2} \\
\midrule
Platform & ISS-mounted & Dedicated satellite \\
Coverage & $\pm 51^\circ$ latitude & $\pm 88^\circ$ latitude \\
Orbit Altitude & $\sim$400\,km & $\sim$500\,km \\
Laser Beams & 3 (2 cover, 1 ref.) & 6 (3 beam pairs) \\
Footprint Diameter & $\sim$25\,m & $\sim$17\,m \\
Along-Track Spacing & $\sim$60\,m & $\sim$0.7\,m \\
Across-Track Spacing & Up to $\sim$600\,m & $\sim$3\,km \\
Vertical Accuracy & $\lesssim$1\,m & $\lesssim$0.1\,m \\
Horizontal Accuracy & $\pm$9\,m & $\pm$6.5\,m \\
Revisit Cycle & ISS orbit-dependent & $\sim$91 days \\
Primary Focus & Forest structure \& biomass & Ice, vegetation, terrain \\
\bottomrule
\end{tabular}
}
\end{table}

\paragraph{Comparison with GEDI.}
Although both GEDI and ICESat-2 provide sparse but precise LiDAR measurements, their coverage and design differ substantially (Table~\ref{tab:gedi_icesat2_specs}). GEDI, hosted on the ISS, is restricted to $\pm 51^\circ$ latitude and emphasizes forest biomass studies, whereas ICESat-2 offers near-global coverage with denser along-track sampling, making it more suitable for validating urban and terrain heights.

\paragraph{Relevant Data Products.}
Two ICESat-2 products are particularly important for calibration:
\begin{itemize}
    \item \textbf{ATL03:} photon-level, precise latitude/longitude/height, with signal-background flags.  
    \item \textbf{ATL08:} terrain and vegetation heights aggregated from ATL03, including ground elevation and canopy metrics.  
\end{itemize}

Despite their sparsity, these products serve as high-fidelity \emph{control points}, enabling systematic correction of dense but uncertain predictions from MHE/MDE models.

\section{Proposed Calibration Pipeline}
\label{sec:icesat2_calibration}
\begin{figure*}[h]
    \centering
    \includegraphics[width=\textwidth]{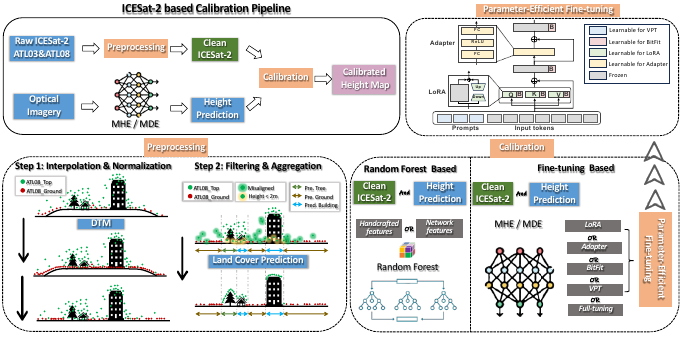}
    \caption{Overview of the proposed ICESat-2 based calibration pipeline. Raw photon data (ATL03/ATL08) are pre-processed and combined with MHE predictions directly, or with MDE predictions after a simple linear fitting against ICESat-2 tracks. 
Calibration is then performed using Random Forest or fine-tuning approaches, yielding a refined and spatially consistent height map.}
    \label{fig:calibration_workflow}
\end{figure*}

This section explains how NASA's ICESat-2 mission data are utilized to calibrate the height predictions obtained from MHE and MDE models. Figure~\ref{fig:calibration_workflow} illustrates the overall pipeline. Optical imagery is first processed through either an MHE or MDE model to generate an initial height map to be calibrated. Raw photon data (ATL03/ATL08) are pre-processed in two main stages: 
(i) DTM-based ground interpolation and normalization, which provide an initial terrain reference, and 
(ii) land-cover–aware filtering and aggregation, which remove spurious returns and consolidate valid samples into a clean subset of ICESat-2 measurements.

For MDE outputs, a simple linear fitting is performed against the clean ICESat-2 tracks to transform relative depth values into absolute heights:  
\begin{equation}
\mathbf{H}_{\mathrm{abs}} = a \cdot \mathbf{D}_{\mathrm{rel}} + b,
\label{eq:linear}
\end{equation}
where $\mathbf{D}_{\mathrm{rel}}$ denotes the relative depth predicted by the MDE, and $\mathbf{H}_{\mathrm{abs}}$ is the corresponding absolute height after calibration. Unless otherwise specified, all subsequent references to original MDE predictions represent these linearly fitted results.

We consider two families of calibration strategies, encompassing a total of seven benchmark methods:  
(1) a Random Forest–based regression approach, using either handcrafted features (\emph{Handcrafted Random Forest, HRF}) or network features (\emph{Network Random Forest, NRF}); and  
(2) parameter-efficient fine-tuning approaches, including LoRA~\citep{hu2022lora}, Adapter~\citep{chen2022adaptformer}, BitFit~\citep{zaken2021bitfit}, and VPT~\citep{jia2022visual}, together with full fine-tuning.  

In both cases, the residuals between the clean ICESat-2 measurements and the initial model predictions along satellite tracks are used as the primary signal for correction. Finally, the selected strategy is applied across the region of interest to yield a refined, spatially consistent height map.

\subsection{ICESat-2 Preprocessing}
\label{subsec:icesat2_preprocessing}

To ensure reliable supervision for height calibration, raw ICESat-2 photon data (ATL03/ATL08) are processed through a streamlined two-step pipeline (as illustrated in the ``Preprocessing'' panels of Figure~\ref{fig:calibration_workflow}). The objective is to derive a clean and consistent subset of above-ground heights that can be directly compared against model predictions.

\paragraph{Step 1: Normalization with Ground Reference.}
We first retain only medium- and high-confidence ATL03 photons (\texttt{signal\_conf} $=3,4$) and those classified by ATL08 as \emph{ground} or \emph{top-of-canopy}. Each photon height is normalized relative to the local ground surface to obtain a \emph{normalized Digital Surface Model (nDSM)}:
\begin{equation}
\mathbf{H}_{\text{nDSM}}(x,y) = \mathbf{H}_{\text{photon}}(x,y) - \mathbf{H}_{\text{ground}}(x,y),
\end{equation}
where $\mathbf{H}_{\text{photon}}$ is the raw photon elevation and $\mathbf{H}_{\text{ground}}$ is the estimated terrain surface.  

\par To estimate $\mathbf{H}_{\mathrm{ground}}$, we employ a two-stage strategy:  
(i) interpolate along-track ground heights using inverse distance weighting (IDW)~\citep{shepard1968two}, and  
(ii) adjust the interpolated profile against FABDEM v1.2\footnote{https://research-information.bris.ac.uk/en/datasets/fabdem-v1-2}, a global $30$\,m-resolution bare-earth \emph{Digital Terrain Model (DTM)}, to suppress large deviations and enforce terrain consistency.  
All non-ground photons are then converted to absolute above-ground heights, while ground photons are fixed at $0$\,m.

\paragraph{Step 2: Land-Cover–Aware Filtering and Aggregation.}
Although ATL08 provides basic photon classification, its accuracy is limited. To refine further, we cross-check each photon against a land-cover product predicted by a segmentation model trained on the OpenEarthMap dataset~\citep{xia2023openearthmap}. Only photons with consistent labels (e.g., both ATL08 and the land-cover model identifying ``tree'' or ``building'') are retained. Implausible outliers are discarded, and remaining non-ground photons are clustered using Density-Based Spatial Clustering of Applications with Noise (DBSCAN)~\citep{ester1996density}. 
It groups nearby photons into stable clusters based on density: 
\begin{equation}
C = \{\,p_i \in P \;\mid\; \mathrm{reachability}(p_i, \varepsilon, \text{MinPts}) = \text{True}\,\},
\end{equation}
where $P$ is the photon set, $\varepsilon$ is the neighborhood radius, and $\text{MinPts}$ is the minimum number of points required to form a cluster. 
Cluster centroids are then used to represent reliable canopy or building heights at the grid-cell level, ensuring robustness against noise and sparsity.

\begin{figure*}[ht!]
    \centering
    \includegraphics[width=\textwidth]{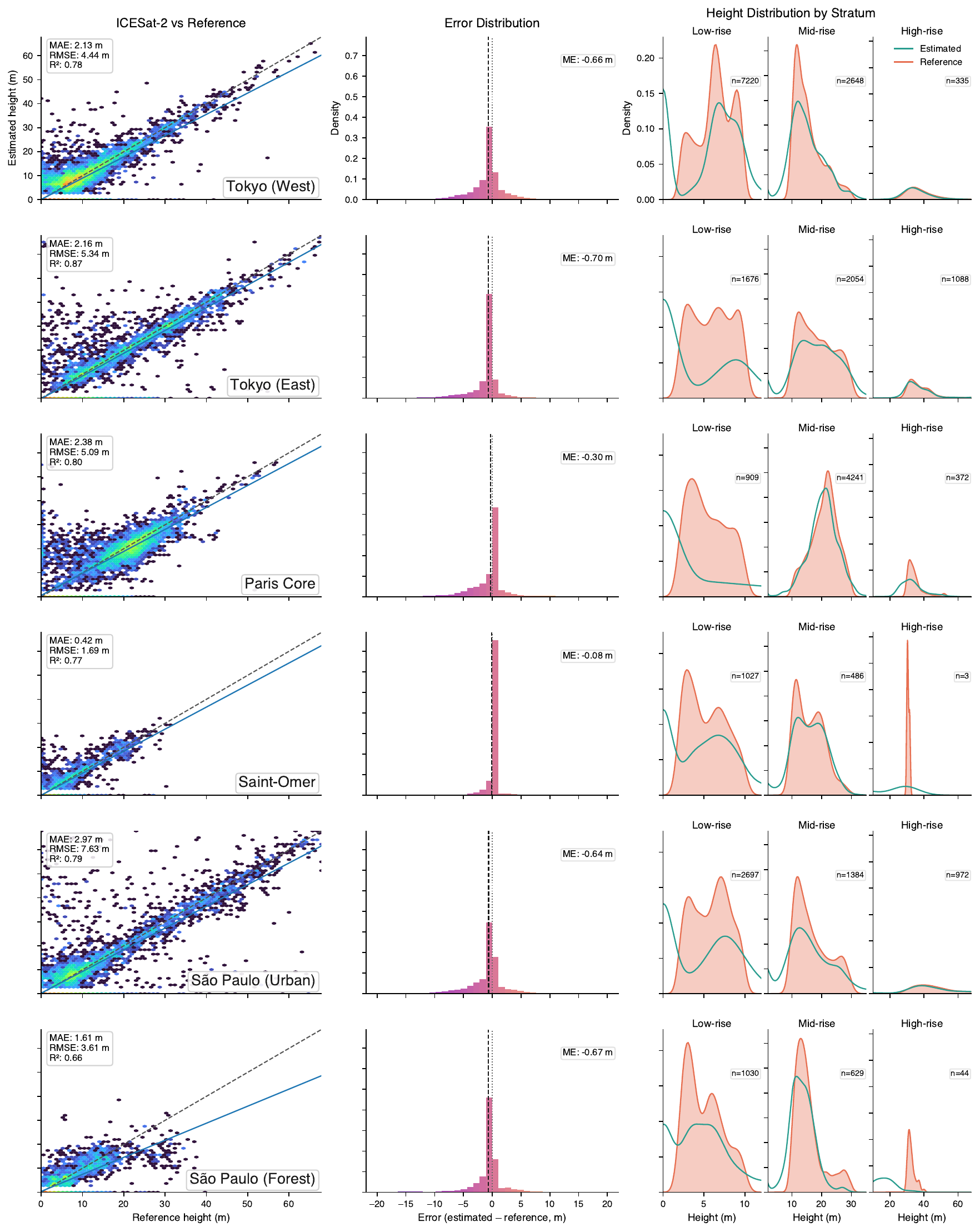}
    \caption{Evaluation of pre-processed ICESat-2 data across six regions. 
For each region, scatter plots compare estimated and reference heights (left), histograms show error distributions (middle), 
and density plots present height distributions for low-, mid-, and high-rise strata (right).}
    \label{fig:vis_preprocessing}
\end{figure*}
\paragraph{Outcome.}
The resulting cleaned photon set (Figure~\ref{fig:vis_preprocessing}) shows markedly reduced scatter and improved agreement with reference nDSMs. Across the six study regions, scatter plots confirm tighter alignment with the one-to-one line, histograms reveal concentrated residuals around zero, and stratified density plots indicate consistent recovery of low-, mid-, and high-rise distributions. The processed dataset exhibits low MAE and RMSE, effectively suppresses spurious noise, and provides a robust, spatially coherent supervisory signal for residual-based calibration.

\subsection{Calibration Strategies}
\label{subsec:calibration}

Once a clean ICESat-2 photon dataset has been obtained (Section~\ref{subsec:icesat2_preprocessing}), we evaluate two families of calibration strategies: (1) Random-Forest–based regression, which adds an external residual learner without modifying model weights; and (2) parameter-efficient fine-tuning, which adapts internal network parameters to learn residuals. In total, seven methods are benchmarked, as illustrated in Figure~\ref{fig:calibration_workflow}.

\paragraph{Residual Definition.}
For each ICESat-2 location $(x_i,y_i)$, let $\mathbf{H}_{\mathrm{pred}}(x_i,y_i)$ be the height predicted by the MDE/MHE model and $\mathbf{H}_{\mathrm{photon}}(x_i,y_i)$ the corresponding cleaned ICESat-2 measurement. We define the residual
\begin{equation}
    \mathbf{r}_i \;=\; \mathbf{H}_{\mathrm{pred}}(x_i,y_i) \;-\; \mathbf{H}_{\mathrm{photon}}(x_i,y_i),
    \label{eq:residual_definition}
\end{equation}
which serves as the supervision target for all calibration methods. At inference, a dense residual field $\widehat{\mathbf{r}}(x,y)$ is estimated and subtracted from the raw prediction:
\begin{equation}
    \mathbf{H}_{\mathrm{corr}}(x,y) \;=\; \mathbf{H}_{\mathrm{pred}}(x,y) \;-\; \widehat{\mathbf{r}}(x,y).
    \label{eq:corr_height_map}
\end{equation}

\subsubsection{Random-Forest–Based Calibration}

\paragraph{Handcrafted-Feature Random Forest (HRF).}
For each photon, we extract a $64\times 64$ window centered at $(x_i,y_i)$ and compute $\sim$27 handcrafted features in four groups:
(i) spatial statistics of the predicted nDSM (mean, std, min, max, 90th, 10th percentiles);
(ii) gradient features from Sobel magnitude (mean, std, 95th percentile);
(iii) optical features from RGB (per-channel mean/std and simple indices such as $(G{-}R)/(G{+}R)$);
(iv) land-cover features (fractions of eight classes and Shannon entropy).
Let $\mathbf{F}^{\text{HRF}}_i$ denote the feature vector for photon $i$. A Random Forest regressor $g(\cdot)$ is trained to predict residuals,
\begin{equation}
    \widehat{\mathbf{r}}(x,y) \;=\; g\!\bigl(\mathbf{F}^{\text{HRF}}_{(x,y)}\bigr),
\end{equation}
which are then applied in Eq.~\eqref{eq:corr_height_map} to produce $\mathbf{H}_{\mathrm{corr}}$ via a sliding-window pass over the image.

\paragraph{Network-Feature Random Forest (NRF).}
Instead of handcrafted features, NRF uses encoder embeddings. For each photon, we retrieve a $d$-dimensional patch embedding $\mathbf{F}^{\text{NRF}}_i \in \mathbb{R}^{1024}$ (e.g., from the first encoder block of RS3DAda or the final encoder block of Depth Anything V2). Training and inference mirror HRF:

\begin{equation}
\begin{split}
    \widehat{\mathbf{r}}(x,y) \;=\; g\!\bigl(\mathbf{F}^{\text{NRF}}_{(x,y)}\bigr), \\
    \mathbf{H}_{\mathrm{corr}}(x,y) \;=\; \mathbf{H}_{\mathrm{pred}}(x,y)
    - \widehat{\mathbf{r}}(x,y).
\end{split}
\end{equation}

\subsubsection{Fine-Tuning–Based Calibration}

Unlike RF-based strategies, these methods adapt internal parameters to \emph{predict} residuals at ICESat-2 locations. Let the network output a residual estimate for each photon location $(x_i,y_i)$ as follows:

\begin{equation}
\widehat{\mathbf{r}}_i ;=; \phi(\mathbf{I}, x_i, y_i;,\Theta)
\end{equation}
where $\mathbf{I}$ denotes the input optical image, $(x_i, y_i)$ represents the coordinates of the $i$-th photon from pre-processed ICESat-2, and $\Theta$ are the learnable network parameters. Given the target $\mathbf{r}_i$ in Eq.~\eqref{eq:residual_definition}, we minimize the Smooth L1 loss
\begin{equation}
    \mathcal{L}(\theta) \;=\; \frac{1}{N} \sum_{i=1}^{N} \mathrm{SmoothL1}\!\bigl(\widehat{\mathbf{r}}_i - \mathbf{r}_i\bigr),
    \label{eq:ft_loss}
\end{equation}
and then deploy the dense residual prediction $\widehat{\mathbf{r}}(x,y)$ in Eq.~\eqref{eq:corr_height_map}.
We benchmark five representative parameter-efficient fine-tuning (PEFT) approaches alongside full fine-tuning:
\begin{itemize}
    \item \textbf{VPT (Visual Prompt Tuning)}~\citep{jia2022visual}: introduces $k$ learnable prompt tokens $\{\mathbf{p}_1,\dots,\mathbf{p}_k\}$ concatenated with the input patch tokens at the Transformer embedding layer. These prompts are updated during training while the backbone remains frozen, effectively steering the encoder’s representation space.
    
    \item \textbf{BitFit}~\citep{zaken2021bitfit}: fine-tunes only the bias terms $\mathbf{b}^{(l)}$ in each Transformer layer. Each block keeps its weight matrices $\mathbf{W}^{(l)}$ frozen, with the residual update governed solely by trainable biases: $\mathbf{h}^{(l)} = \mathbf{W}^{(l)}\mathbf{x}^{(l)} + \mathbf{b}^{(l)}$.
    
    \item \textbf{LoRA}~\citep{hu2022lora}: augments the attention projection matrices (e.g., $\mathbf{W}_q, \mathbf{W}_v$) with a low-rank decomposition $\mathbf{A}\mathbf{B}^\top$, where $\mathbf{A} \in \mathbb{R}^{d \times r}, \mathbf{B} \in \mathbb{R}^{r \times d}$ and $r \ll d$. The effective weight becomes $\mathbf{W} = \mathbf{W}_0 + \mathbf{A}\mathbf{B}^\top$, enabling efficient adaptation with minimal trainable parameters.
    
    \item \textbf{Adapter}~\citep{chen2022adaptformer}: inserts lightweight bottleneck modules after the feed-forward network (FFN) in each Transformer block. An adapter consists of a down-projection $\mathbf{W}_{\downarrow}$, nonlinearity $\sigma(\cdot)$, and up-projection $\mathbf{W}_{\uparrow}$: $f_{\text{adapter}}(x) = \mathbf{W}_{\uparrow}\sigma(\mathbf{W}_{\downarrow}x)$. The block output becomes $x_{\text{out}} = x + \mathrm{FFN}(x) + f_{\text{adapter}}(x)$, with the backbone weights frozen.

    \item \textbf{Full Fine-Tuning}: update all parameters end-to-end.
\end{itemize}

\paragraph{Practical Notes and Summary.}
Training samples are drawn from patches intersected by ICESat-2 tracks, using residuals $\mathbf{r}_i$ as supervision; at test time, the learned residual field $\widehat{\mathbf{r}}(x,y)$ is predicted densely over the scene. RF-based methods (HRF, NRF) are lightweight and do not alter the backbone, making them efficient and easily transferable across regions. Fine-tuning–based methods directly adapt the network to the local domain, often achieving stronger gains at higher computational and storage costs. Together, these seven strategies provide a comprehensive, unified benchmark for ICESat-2–guided height calibration.

\section{Experiments}
\label{sec:benchexperiments}
\begin{table*}[ht!]
    \centering
    \caption{PEFT configurations used in our benchmark. All methods share the same optimizer, schedule, and data protocol described in the text.}
    \label{tab:peft_config}
    \adjustbox{max width=\textwidth}{
    \begin{tabular}{l l l}
        \toprule
        \textbf{Method} & \textbf{Trainable components} & \textbf{Key hyperparameters} \\
        \midrule
        VPT~\citep{jia2022visual} 
        & $k$ learnable prompt tokens at ViT input 
        & $k{=}5$ prompts \\
        BitFit~\citep{zaken2021bitfit} 
        & Bias terms in all Transformer layers 
        & biases only \\
        LoRA~\citep{hu2022lora} 
        & Low-rank updates on attention projections (e.g., $W_q,W_v$) 
        & rank $r{=}4$ \\
        Adapter (AdaptFormer)~\citep{chen2022adaptformer} 
        & Bottleneck modules after FFN in each block 
        & bottleneck $r{=}16$ \\
        Full fine-tuning 
        & All parameters end-to-end 
        & same LR/schedule as PEFT \\
        \bottomrule
    \end{tabular}}
\end{table*}
Having established our sparse-data correction method (Section~\ref{sec:icesat2_calibration}), we now evaluate its effectiveness across six diverse study areas spanning Europe, Asia, and South America. These areas encompass a broad spectrum of urban and peri-urban forms: dense historic city cores (Paris, France), mid-sized towns with mixed residential and agricultural surroundings (Saint-Omer, France), heterogeneous metropolitan districts (Tokyo East and Tokyo West, Japan), a dense high-rise megacity core (S\~ao Paulo Urban, Brazil), and sparsely inhabited mountainous forest zones (S\~ao Paulo Forest, Brazil). 

This section presents both quantitative and qualitative analyses of the corrected nDSM results, highlighting the robustness of our pipeline under varying land-cover conditions, building typologies, and development densities—from compact high-rise clusters to suburban neighborhoods and tree-covered rural landscapes.

\subsection{Experimental Setup}
\label{sec:exp_setup}

\paragraph{Random Forest.}
We include two RF-based residual learners: 
\emph{HRF} uses $64{\times}64$ image patches with a 27-D handcrafted feature vector per patch and a forest of 100 trees; 
\emph{NRF} uses $14{\times}14$ ViT-L encoder embeddings (1024-D) at photon locations with 100 trees.

\paragraph{PEFT and full fine-tuning.}
All fine-tuning methods are trained to predict residuals (Section~\ref{subsec:calibration}); the prediction head is unfrozen in every case. 
Unless otherwise noted, we use \textbf{AdamW} (learning rate $5{\times}10^{-4}$, weight decay $1{\times}10^{-4}$, warmup 5 epochs), batch size 2, dropout $0.1$, and early stopping (patience 8, min\_delta 0.25). 
Training is performed on an NVIDIA A100 GPU. 
Input crop sizes are $392{\times}392$ for \textbf{RS3DAda} and $518{\times}518$ for \textbf{Depth Anything V2}. 
Full fine-tuning follows the same schedule as PEFT. The detailed hyperparameter settings for all PEFT methods are summarized in Table~\ref{tab:peft_config}.

\subsection{Evaluation Metrics}
\label{subsec:evaluation_metrics}

To comprehensively assess the accuracy and structural fidelity of corrected height maps, we employ four complementary metrics: Mean Absolute Error (MAE), Root Mean Squared Error (RMSE), Structural Similarity Index (SSIM)~\citep{wang2004image}, and the F1 Score for Height Estimation ($F_{1}^{HE}$)~\citep{song2024synrs3d}. Together, they evaluate not only overall error magnitudes but also structural consistency and the reliability of predictions for significant above-ground objects.

\paragraph{Mean Absolute Error (MAE).}
MAE measures the average absolute deviation between predicted heights $\hat{\mathbf{Y}}$ and reference heights $\mathbf{Y}$:
\begin{equation}
    \mathrm{MAE} = \frac{1}{N} \sum_{i=1}^{N} \bigl|\hat{\mathbf{Y}}_i - \mathbf{Y}_i \bigr|.
\end{equation}
It emphasizes overall accuracy by penalizing each error equally, making it robust to outliers.

\paragraph{Root Mean Squared Error (RMSE).}
RMSE penalizes larger errors more strongly:
\begin{equation}
    \mathrm{RMSE} = \sqrt{\frac{1}{N} \sum_{i=1}^{N} \bigl(\hat{\mathbf{Y}}_i - \mathbf{Y}_i \bigr)^2}.
\end{equation}
This metric is sensitive to large deviations, highlighting the presence of significant height mismatches.

\paragraph{Structural Similarity Index (SSIM).}
SSIM evaluates the structural similarity between two images by jointly considering luminance, contrast, and structural components:
\begin{equation}
    \mathrm{SSIM}(\mathbf{\mathbf{Y}}, \hat{\mathbf{Y}}) = 
    \frac{(2\mu_Y \mu_{\hat{\mathbf{Y}}} + C_1)(2\sigma_{Y\hat{Y}} + C_2)}{(\mu_Y^2 + \mu_{\hat{\mathbf{Y}}}^2 + C_1)(\sigma_Y^2 + \sigma_{\hat{\mathbf{Y}}}^2 + C_2)},
\end{equation}
where $\mu$, $\sigma^2$, and $\sigma_{\mathbf{Y}\hat{\mathbf{Y}}}$ denote means, variances, and covariance of patches, and $C_1, C_2$ are small constants for numerical stability. SSIM complements MAE/RMSE by focusing on structural fidelity, ensuring that edges and fine details are preserved.

\paragraph{F1 Score for Height Estimation ($\mathrm{F}_{1}^{\mathrm{HE}}$).}
The $\mathrm{F}_{1}^{\mathrm{HE}}$ score adapts the traditional F1 metric to height estimation, emphasizing precision and recall for objects above a threshold $T$ (e.g., $1$\,m). A prediction is considered correct if its relative error $\delta$ is within a tolerance $\eta$:
\begin{align}
    \mathrm{TP} &= \sum\left( \left(\hat{\mathbf{Y}} > \text{T} \land \mathbf{Y} > \text{T}\right) \land \left(\delta < \eta \right) \right), \\
    \mathrm{FP} &= \sum\left( \hat{\mathbf{Y}} > \text{T} \land \mathbf{Y} \leq \text{T} \right), \\
\mathrm{FN} &= \sum\left( \hat{\mathbf{Y}} \leq \text{T} \land \mathbf{Y} > \text{T} \right).
\end{align}
From these, precision, recall, and $\text{F}_{1}^{\text{HE}}$ are defined as:
\begin{align}
    \text{Precision} &= \frac{\text{TP}}{\text{TP} + \text{FP}}, \quad
    \text{Recall} = \frac{\text{TP}}{\text{TP} + \text{FN}}, \\
    \text{F}_{1}^{\text{HE}} &= 2 \times \frac{\text{Precision} \times \text{Recall}}{\text{Precision} + \text{Recall}}.
\end{align}
Unlike MAE or RMSE, $F_{1}^{HE}$ specifically targets the accurate detection of non-ground objects (buildings, trees), ensuring that the evaluation reflects practical relevance in urban and forested environments.

\paragraph{Summary.}
MAE and RMSE provide global error magnitudes, SSIM measures structural fidelity, and $F_{1}^{HE}$ emphasizes correctness for above-ground structures. This combination ensures that our evaluation captures both average error reduction and improvements in the geometric consistency of urban/forest form.

\subsection{Experimental Results and Analysis}
\label{sec:experiments}
To comprehensively evaluate the proposed ICESat-2 calibration pipeline, we present detailed results and analyses from multiple perspectives. 
We begin with overall performance, demonstrating the average improvements of all calibration methods compared with baselines. 
We then examine generalization across diverse geographic environments to assess adaptability. 
Next, we analyze several key phenomena observed in the experiments, providing deeper insights into model behavior. 
Finally, we discuss practical trade-offs between accuracy, efficiency, and resource consumption, leading to method recommendations and qualitative validations.

\subsubsection{Overall Performance Comparison}
\begin{figure*}[ht!]
    \centering
    \includegraphics[width=\textwidth]{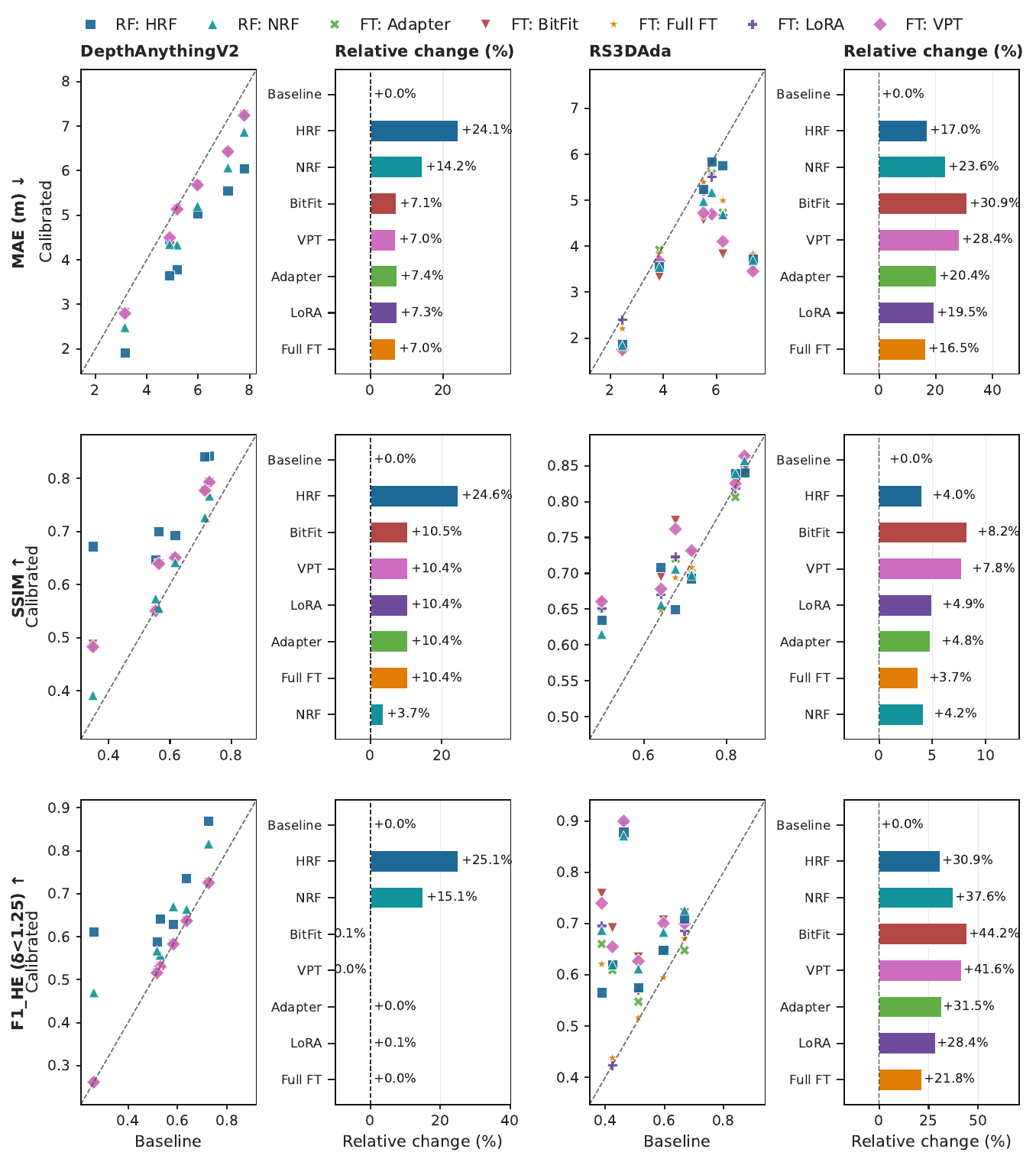}
    \caption{Calibration results on Depth Anything V2 and RS3DAda across six regions. Scatter plots show baseline vs. calibrated metrics (MAE, SSIM, $F_1^{\text{HE}}$) for each region and method. Bar plots summarize the averaged relative improvements across all six regions for Random Forest and fine-tuning methods.}
    \label{fig:calibration_composite_panel}
\end{figure*}
Figure~\ref{fig:calibration_composite_panel} presents the calibration results across six study regions for both Depth Anything V2 (MDE) and RS3DAda (MHE). Scatter plots (left) show baseline versus calibrated outputs for each region and method, while bar plots (right) summarize the averaged relative improvements across all six regions in terms of the three evaluation metrics (MAE, SSIM, $F_{1}^{HE}$).

Our results clearly demonstrate that all seven calibration methods consistently outperform the baseline in most cases, confirming the general effectiveness of the proposed ICESat-2 calibration pipeline. 
As shown in the scatter plots of Figure~\ref{fig:calibration_composite_panel}, for MAE (where lower is better), most calibration points lie \emph{below} the diagonal, while for SSIM and $F_{1}^{HE}$ (where higher is better), the points lie \emph{above} the diagonal. 
This pattern highlights the broad benefits of residual correction with ICESat-2 guidance across different evaluation perspectives.
The bar plots further quantify these gains, showing that calibration delivers substantial relative improvements.

A closer inspection reveals distinct preferences between the two base models. 
For the remote sensing–specific RS3DAda, fine-tuning approaches achieve the largest gains, with lightweight methods such as BitFit and VPT leading to the most notable improvements across all three metrics. 
Random Forest–based strategies also show stable benefits, though to a slightly lesser extent. 
In contrast, for Depth Anything V2 model, Random Forest–based methods (HRF, NRF) dominate, achieving up to $+24.1\%$ reduction in MAE and $+24.6\%$ improvement in SSIM. 
By comparison, fine-tuning methods provide only modest improvements. 
This divergence suggests that the alignment between the pretraining task and the target calibration task plays a crucial role in determining which family of methods is most effective—a point we analyze in more detail in Section~\ref{subsec:insights}.

\subsubsection{Performance Across Diverse Geographic Environments}
\label{subsec:geographic_performance}
\begin{figure*}[ht!]
    \centering
    \includegraphics[width=\textwidth]{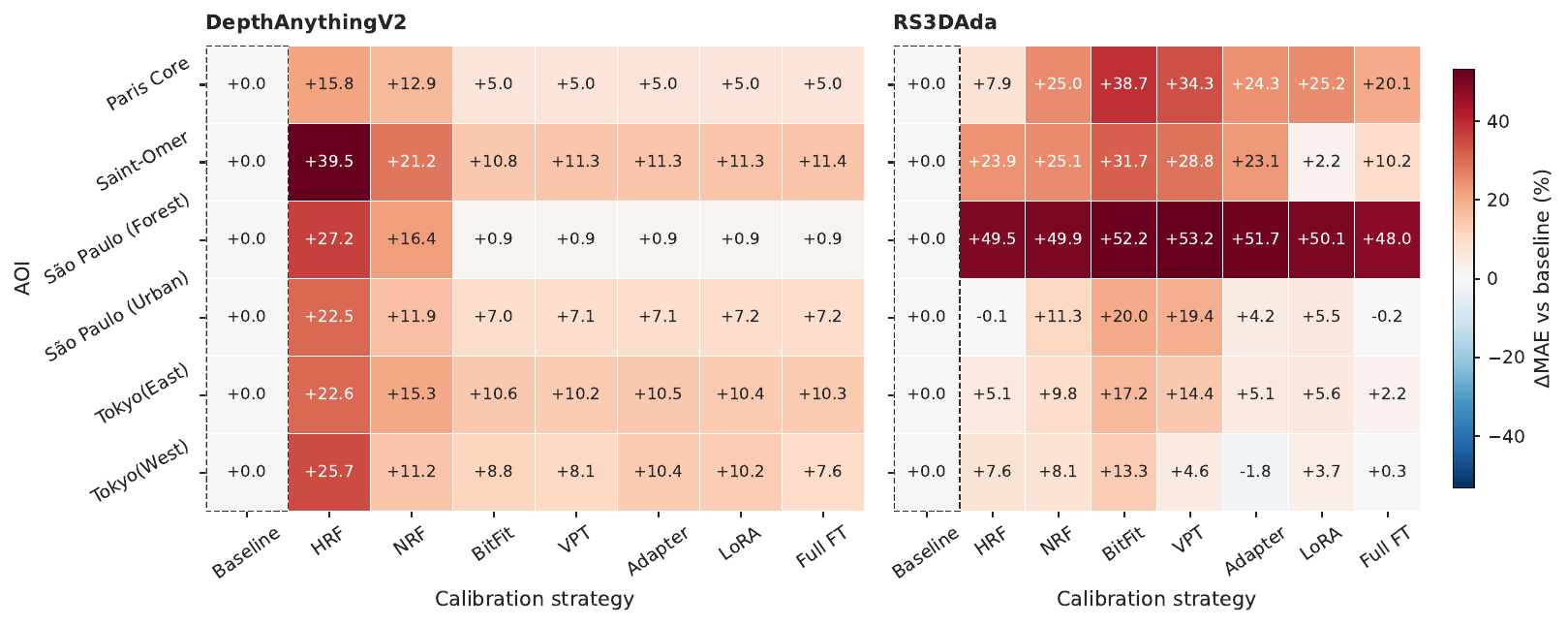}
    \caption{Relative MAE improvements (\%) of different calibration strategies across six regions for Depth Anything V2 and RS3DAda.}
    \label{fig:heatmap}
\end{figure*}
To assess the robustness and generalization of calibration strategies, we evaluate all methods across six heterogeneous regions spanning nearly 300~km$^2$, including dense urban cores, peri-urban towns, and forested landscapes. Figure~\ref{fig:heatmap} presents a heatmap of relative MAE improvements (\%) for each calibration strategy in each region.

For the RS3DAda model, the heatmap reveals that nearly all calibration strategies deliver consistent performance gains across regions. A particularly striking result emerges in the \emph{São Paulo Forest} area, where almost every strategy achieves close to $50\%$ improvement in MAE. This suggests that the model, originally trained on synthetic datasets dominated by urban scenes, lacked sufficient exposure to pure forest environments. Sparse but reliable ICESat-2 observations thus play a critical role in compensating for this weakness and greatly enhance performance in such regions. Moreover, as shown in Figure~\ref{fig:area1}, areas with relatively small std in above-ground object heights (e.g., São Paulo Forest, Paris Core, and Saint-Omer) exhibit substantial improvements across nearly all calibration strategies. By contrast, regions with larger std remain challenging, and the gains are comparatively limited, indicating a strong correlation between calibration effectiveness and the underlying std of above-ground objects.

In contrast, Depth Anything V2 exhibits more modest gains in forest areas. While Random Forest–based approaches (HRF and NRF) consistently outperform other methods across all six regions, parameter-efficient fine-tuning shows only limited benefits. This pattern aligns closely with the overall results in Figure~\ref{fig:calibration_composite_panel}, reinforcing the observation that RF-based methods are more effective when adapting models like Depth Anything V2, whose pretraining emphasizes relative depth from natural images rather than absolute height in remote sensing contexts.

\subsubsection{In-depth Analysis of RF-based Calibration}
\label{subsec:rfanylysis}

To gain deeper insight into how the RF corrector performs height calibration and to inform its practical deployment, we conducted two complementary studies: a grouped feature-importance analysis and a hyperparameter-sensitivity analysis.

\begin{figure}[t!]
\centering
\includegraphics[width=\linewidth]{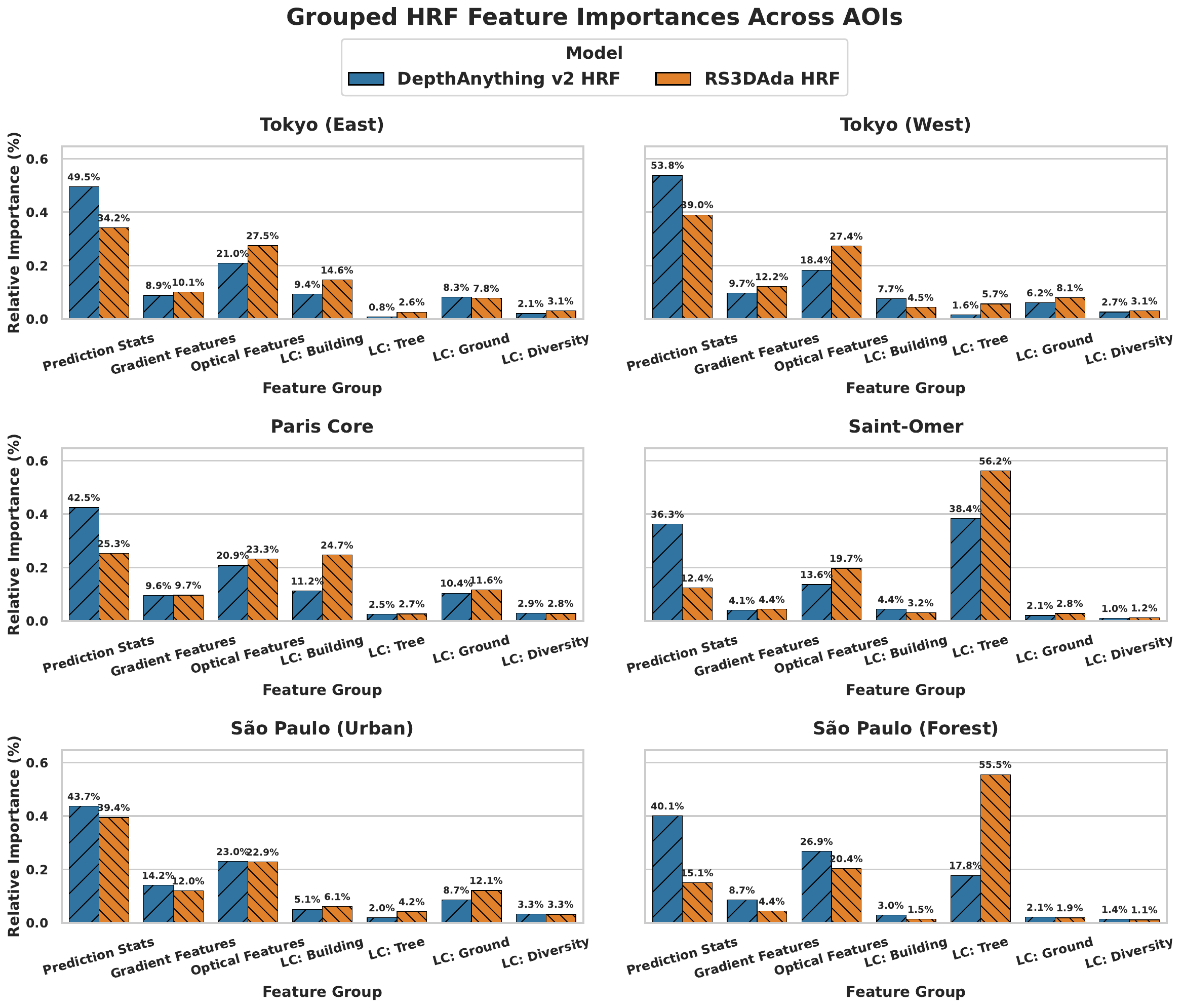}
\caption{Grouped feature-importance analysis for the HRF corrector across all models and AOIs. 
\textit{Prediction Stats} dominate in urban regions (e.g., Tokyo, São~Paulo Urban). 
\textit{LC:~Tree} is most critical in forested areas (e.g., São~Paulo Forest). 
\textit{Optical Features} and \textit{LC:~Building} show significant, area-specific contributions, whereas \textit{Gradient Features} generally exhibit slightly lower contributions.}
\label{fig:feature_importance}
\end{figure}

\paragraph{Feature importance.}
Since NRF relies on features extracted from a neural encoder, we report grouped importance only for the handcrafted RF (HRF), where each feature is explicit and interpretable. 
The 27 features are organized into various groups: 
\textit{Prediction Stats} (mean, std, min, max, p10, p90 of the predicted height in a patch), 
\textit{Gradient Features} (mean, std, p95 of gradient magnitude derived from the predicted height), 
\textit{Optical Features} (RGB means and stds, NDVI-like mean and std, red–green ratio), 
and \textit{Land-Cover (LC)} descriptors (fractions of eight LC classes and Shannon diversity).
Figure~\ref{fig:feature_importance} summarizes results across six AOIs for both backbones. 
The analysis reveals strong context dependence: 
1)~\textit{Prediction Stats} dominate in dense urban regions such as Tokyo and São~Paulo (Urban), with additional area-specific contributions from \textit{Optical Features} and \textit{LC:~Building}. 
2)~In heterogeneous or forested areas such as Saint-Omer and São~Paulo (Forest), \textit{LC:~Tree} becomes the most critical cue. 
3)~Compared with RS3DAda, Depth Anything V2 relies more on \textit{Prediction Stats}, whereas RS3DAda benefits more from semantic cues. 
This contrast reflects their distinct pretraining objectives: 
Depth Anything V2, trained for relative depth on natural images, retains scale-related biases effectively corrected by statistical descriptors. 
RS3DAda, pretrained for absolute height on synthetic remote-sensing imagery, already learns global scale but exhibits domain gaps in object semantics and surface context, 
making semantic cues more informative.

\begin{figure}[t!]
\centering
\includegraphics[width=\linewidth]{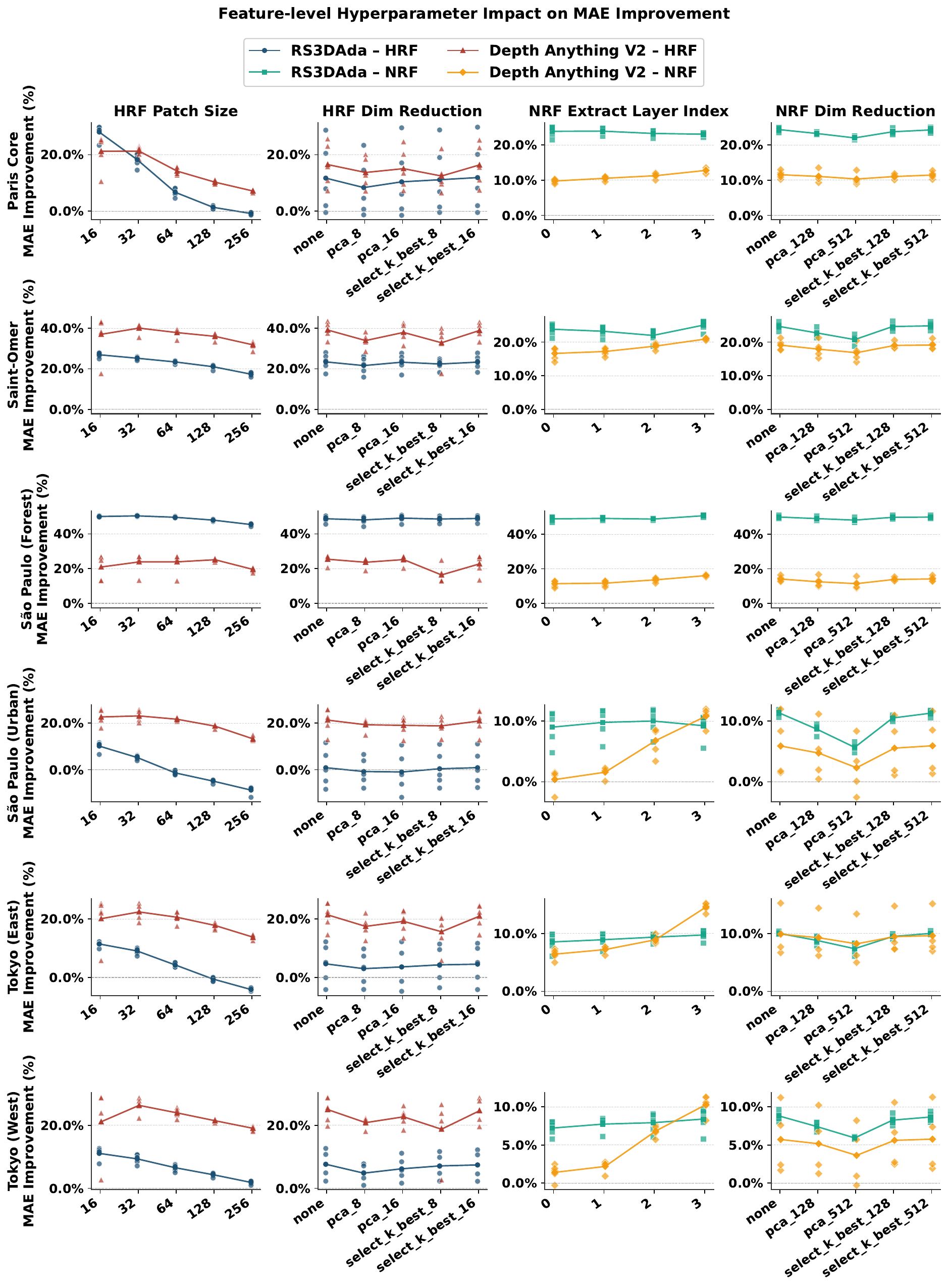}
\caption{Sensitivity of feature-level parameters across six AOIs. 
Four parameter groups are evaluated: HRF input patch size, HRF feature-compression dimension, NRF encoder-layer index, and NRF feature-compression dimension.}
\label{fig:feature_hyperparam}
\end{figure}

\paragraph{Sensitivity to hyperparameters.}
We categorize tunable factors into two classes: 
\textit{RF-internal hyperparameters} (number of trees, maximum depth, maximum features, and minimum samples per leaf) and 
\textit{feature-level hyperparameters} that control the input and representation. 
The latter include, for HRF, the input patch size and feature-compression scheme, and for NRF, the encoder layer index and compression scheme. 
Our experiments show that results are robust to RF-internal parameters; hence, for clarity, we omit their plots and focus on the more sensitive \emph{feature-level} hyperparameters. 
Figure~\ref{fig:feature_hyperparam} illustrates that smaller HRF patch sizes consistently yield higher MAE improvement than larger ones. 
However, runtime increases rapidly as patch size decreases; for instance, a size of 16 takes over ten times longer than 64. 
We therefore adopt 64 as a balanced choice between accuracy and efficiency, which is consistently applied in the main experiments and reported tables. Feature-compression parameters show only a minor impact within the tested range; neither \textit{PCA} (Principal Component Analysis) nor \textit{select\_k\_best} provided consistent gains, and the uncompressed setting (\textit{none}) achieved the best overall performance. 
For NRF, RS3DAda is largely insensitive to encoder-layer depth, while Depth Anything V2 benefits from deeper layers with richer semantics. 
This observation complements the HRF feature importance findings: feature importance reveals \emph{what} cues each model relies on, whereas NRF sensitivity identifies \emph{where} these cues reside in the backbone. 
RS3DAda, which was pre-trained for \textit{absolute height estimation} in the \textit{remote sensing domain}, is robust and insensitive to layer choice because its entire feature hierarchy is already highly relevant to our \textit{height calibration} task.
Conversely, Depth Anything V2, being domain-mismatched (pre-trained on \textit{relative depth} from \textit{natural images}), is sensitive to layer choice and strongly prefers deep semantic features, as only these abstract concepts are transferable to overhead imagery to fix its domain gap.

Taken together, these results indicate that the effectiveness of RF-based calibration depends primarily on the quality and scale of the input representations rather than fine-grained tuning of the RF itself.

\subsubsection{Key Findings}
\label{subsec:insights}

\noindent\textbf{Insight 1: Why RF Outperforms Fine-Tuning on Depth Anything V2.}
\par A consistent observation is that Random Forest–based calibration achieves larger gains than fine-tuning on Depth Anything V2. This stems from a fundamental \emph{domain gap} between the model’s pretraining task and our calibration objective. Depth Anything V2 was trained on natural images for \emph{relative depth} estimation, meaning its feature space is not naturally aligned with absolute height values in remote sensing imagery. In our pipeline, an initial linear fitting step already performs a global scale calibration of the raw outputs, as shown in Eq.~\eqref{eq:linear}, removing most systematic bias. Random Forest, acting as an external learner, then performs a second-stage \emph{local refinement} of residuals, which plays exactly to its strengths. In contrast, fine-tuning attempts to adapt the entire large-scale network to sparse ICESat-2 absolute heights, which both conflicts with the pretrained knowledge (relative structure rather than absolute scale) and easily leads to overfitting under sparse supervision.

\noindent\textbf{Insight 2: Why Small-Parameter Fine-Tuning Excels on RS3DAda.}
\par
For RS3DAda, the situation is fundamentally different: its pretraining task on synthetic remote sensing data explicitly targets \emph{absolute height estimation}, which is highly consistent with ICESat-2 supervision. Fine-tuning therefore provides an effective mechanism to bridge the synthetic-to-real gap.

Interestingly, our experiments reveal that the strongest improvements arise from the most extreme parameter efficient methods, BitFit and VPT (tuning $<$0.09\% and $<$0.05\% of total parameters, respectively), while gains diminish as the number of trainable parameters increases. For comparison, LoRA ($r{=}4$) and Adapter ($r{=}16$) tune approximately 0.13\% and 0.27\% of the parameters. 
We hypothesize that this minimal intervention is optimally suited for the extremely sparse supervisory signal from ICESat-2, whereas methods tuning more parameters (like LoRA and Adapter) are more prone to overfitting under limited supervision. 
This hypothesis is further supported by our convergence observations: BitFit and VPT converge fastest (approximately 17–20~epochs), while LoRA and Adapter require substantially more iterations (around 30–35~epochs).

Furthermore, we speculate that BitFit's slight performance edge over VPT stems from its specific mechanism: it tunes \emph{only} the bias terms. While RS3DAda already outputs absolute heights (unlike Depth Anything V2), it does not include an explicit linear fitting step (Eq.~\ref{eq:linear}). BitFit's bias-only tuning may implicitly perform a similar role, applying a lightweight scale–offset correction to the output distribution while preserving the integrity of the pretrained feature space. This property likely makes it the most robust and stable strategy for this task.

This finding highlights a crucial principle: when supervision is sparse, as with ICESat-2 track data, limiting the number of trainable parameters is essential to prevent overfitting and to unlock the full potential of fine-tuning for real-world height calibration.

\subsubsection{Practical Trade-offs and Method Selection}
\begin{figure*}[htbp]
    \centering
    \includegraphics[width=\textwidth]{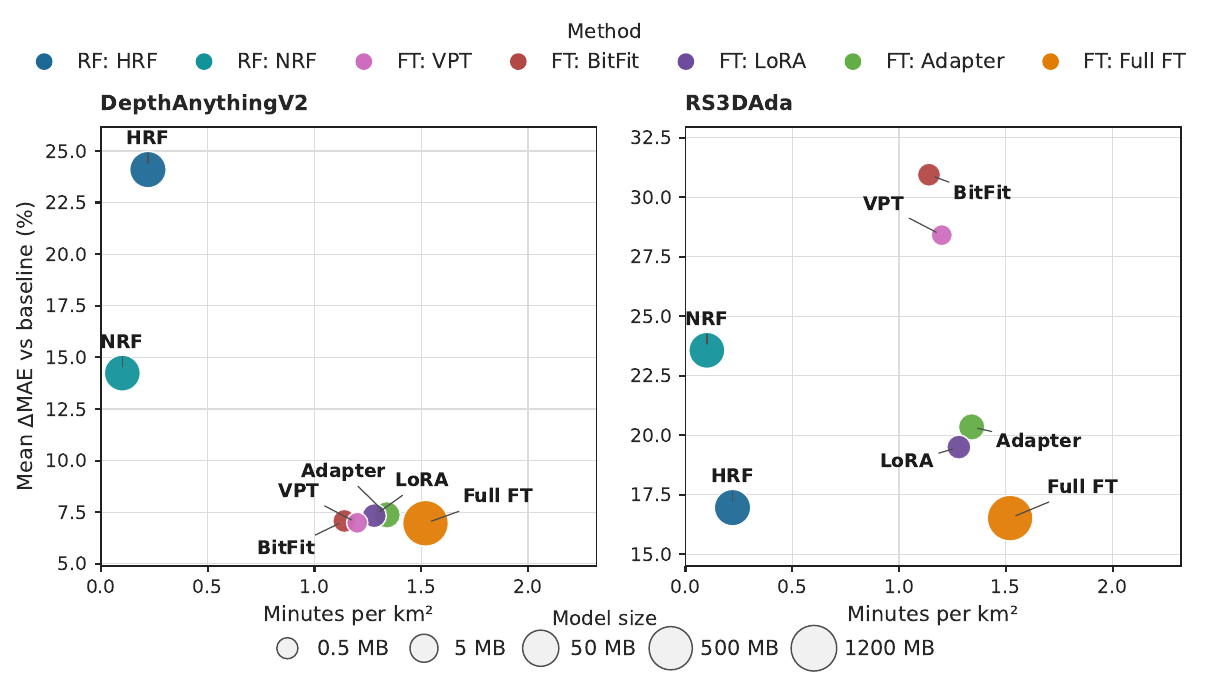}
    \caption{Comparison of fine-tuning and calibration strategies in terms of accuracy gain, calibration time (train + inference), and model size on Depth Anything V2 and RS3DAda. Here, ``model'' size refers specifically to the size of the stored/trainable parameters. 
    }
    \label{fig:trade_off}
\end{figure*}

\begin{table*}[htbp]
\centering
\caption{Computational cost comparison of calibration methods. 
“Model Size” refers to the size of trainable parameters. 
GPU memory usage is reported for both backbones (Depth Anything V2, RS3DAda). 
Calibration time is measured from fine-tuning to inference, normalized per~km$^2$.}
\label{tab:comp_cost}
\sisetup{
  table-format=5.1,
  round-mode=places,
  round-precision=1,
  detect-weight=true,
  detect-inline-weight=math
}
\begin{tabular}{
  l
  l
  S[table-format=5.1]
  S[table-format=4.1]
  S[table-format=4.1]
  S[table-format=4.1]
  S[table-format=4.1]
  S[table-format=1.2]
}
\toprule
\multirow{2}{*}{Category} & \multirow{2}{*}{Method} 
& \multicolumn{2}{c}{GPU Memory (MiB) — DAV2} 
& \multicolumn{2}{c}{GPU Memory (MiB) — RS3DAda} 
& {Model Size} & {Time} \\
\cmidrule(lr){3-4} \cmidrule(lr){5-6}
& & {Train} & {Inference} & {Train} & {Inference} & {(MB)} & {(min/km$^2$)} \\
\midrule
\multirow{2}{*}{RF-Based} 
& HRF{\textsuperscript{\dag}} & 0.0 & 0.0 & 0.0 & 0.0 & 55.1 & 0.22 \\ 
& NRF{\textsuperscript{\ddag}} & 1674.4 & 1674.7 & 1499.5 & 1499.5 & 48.8 & 0.10 \\ 
\midrule
\multirow{5}{*}{PEFT-Based} 
& VPT & 6035.4 & 2062.0 & 4122.4 & 1716.5 & 0.54 & 1.20 \\
& BitFit & 5995.4 & 2061.8 & 4115.1 & 1716.5 & 1.22 & 1.14 \\
& LoRA & 6307.4 & 2063.5 & 4296.9 & 1718.0 & 1.64 & 1.28 \\
& Adapter & 6740.9 & 2065.2 & 4547.0 & 1719.8 & 3.42 & 1.34 \\
& Full~FT & 11056.4 & 2061.8 & 8635.7 & 1716.5 & 1278.0 & 1.52 \\
\midrule
\multicolumn{7}{l}{\textbf{Average Calibration Time Ratio (PEFT / RF-Based)}} 
& \textbf{7.75$\times$} \\
\bottomrule
\end{tabular}

\vspace{0.4em}
\begin{minipage}{0.96\textwidth}\footnotesize
\textbf{Notes.} 
{\dag}~pure CPU method, {\ddag}~uses GPU only for feature extraction.
\end{minipage}
\end{table*}

\paragraph{Trade-off analysis.}

To inform real-world deployment, we analyze the trade-offs between accuracy, calibration time, and model size across all calibration strategies (Figure~\ref{fig:trade_off} and Table~\ref{tab:comp_cost}). 
Both figures summarize results measured under identical experimental settings: batch size~2 for training, batch size~4 for inference, and input image sizes of 518$\times$518 (Depth Anything V2) and 392$\times$392 (RS3DAda). 
For RF-based methods, HRF is a CPU-only approach and NRF uses the GPU solely for feature extraction.

Table~\ref{tab:comp_cost} provides quantitative computational statistics that complement the trade-off landscape in Figure~\ref{fig:trade_off}. 
GPU memory usage is reported for both backbones during training and inference, together with model size (trainable parameters) and normalized runtime per~km$^2$ from fine-tuning to inference. 
As expected, inference memory across PEFT variants remains nearly constant (differences $<$5\%), while calibration time scales moderately with model size. The overall GPU memory usage is higher for Depth Anything V2 than for RS3DAda, primarily due to the larger input resolution (518$\times$518 vs.~392$\times$392), which increases the size of intermediate activation maps during both training and inference.

For \textbf{Depth~Anything~V2}, RF-based strategies, particularly HRF, clearly dominate: they achieve the highest accuracy improvements at minimal cost, with model sizes below~60~MB and runtimes below~0.3~min/km$^2$. 
This makes them highly cost-effective for operational use. 

In contrast, for \textbf{RS3DAda}, the largest gains arise from lightweight fine-tuning strategies, especially BitFit and VPT, which require roughly 7--8$\times$ more calibration time than RF but yield substantially higher accuracy. While these approaches incur higher calibration time, their superior accuracy makes them preferable when precision is paramount. This contrast highlights that optimal method selection depends on both the backbone model and the application requirements.

\paragraph{Summary of best-performing methods.}
\begin{table*}[htbp]
\centering
\caption{Depth Anything V2 baseline vs HRF performance. Numbers in parentheses indicate relative improvements.}
\label{tab:dav2_hrf}
\adjustbox{max width=\textwidth}{
\begin{tabular}{l *{8}{c}}
\toprule
\multirow{2}{*}{AOI} 
  & \multicolumn{2}{c}{MAE $\downarrow$ (m)} 
  & \multicolumn{2}{c}{RMSE $\downarrow$ (m)} 
  & \multicolumn{2}{c}{SSIM $\uparrow$} 
& \multicolumn{2}{c}{$\text{F1}^{\text{HE}}_{\delta<1.25} \uparrow$} \\
\cmidrule(lr){2-3} \cmidrule(lr){4-5} \cmidrule(lr){6-7} \cmidrule(lr){8-9}
  & {Baseline} & {HRF} & {Baseline} & {HRF} & {Baseline} & {HRF} & {Baseline} & {HRF} \\
\midrule
Tokyo(E)     & 7.160 & \textbf{5.540} {\scriptsize (+22.6\%)} & 13.235 & \textbf{11.321} {\scriptsize (+14.5\%)} & 0.729 & \textbf{0.842} {\scriptsize (+15.6\%)} & 0.517 & \textbf{0.588} {\scriptsize (+13.6\%)} \\
Tokyo(W)     & 4.892 & \textbf{3.635} {\scriptsize (+25.7\%)} & 10.275 & \textbf{8.584} {\scriptsize (+16.5\%)} & 0.714 & \textbf{0.840} {\scriptsize (+17.7\%)} & 0.637 & \textbf{0.735} {\scriptsize (+15.5\%)} \\
Paris Core   & 5.976 & \textbf{5.034} {\scriptsize (+15.8\%)} & 7.777 & \textbf{6.850} {\scriptsize (+11.9\%)} & 0.617 & \textbf{0.692} {\scriptsize (+12.1\%)} & 0.583 & \textbf{0.628} {\scriptsize (+7.7\%)} \\
Saint-Omer   & 3.150 & \textbf{1.905} {\scriptsize (+39.5\%)} & 4.902 & \textbf{3.804} {\scriptsize (+22.4\%)} & 0.349 & \textbf{0.671} {\scriptsize (+92.6\%)} & 0.262 & \textbf{0.611} {\scriptsize (+133.3\%)} \\
S\~ao Paulo Urban    & 7.793 & \textbf{6.036} {\scriptsize (+22.5\%)} & 11.266 & \textbf{10.064} {\scriptsize (+10.7\%)} & 0.564 & \textbf{0.699} {\scriptsize (+24.0\%)} & 0.530 & \textbf{0.641} {\scriptsize (+20.8\%)} \\
S\~ao Paulo Forest   & 5.184 & \textbf{3.776} {\scriptsize (+27.2\%)} & 7.257 & \textbf{5.678} {\scriptsize (+21.8\%)} & 0.553 & \textbf{0.647} {\scriptsize (+16.9\%)} & 0.726 & \textbf{0.869} {\scriptsize (+19.8\%)} \\
\midrule
\textbf{Avg} & 5.692 & \textbf{4.321} {\scriptsize (+24.1\%)} & 9.119 & \textbf{7.717} {\scriptsize (+15.4\%)} & 0.588 & \textbf{0.732} {\scriptsize (+24.6\%)} & 0.542 & \textbf{0.679} {\scriptsize (+25.1\%)} \\
\bottomrule
\end{tabular}
}
\end{table*}

\begin{table*}[htbp]
\centering
\caption{RS3DAda baseline vs BitFit performance. Numbers in parentheses indicate relative improvements.}
\label{tab:rs3dada_bitfit}
\adjustbox{max width=\textwidth}{
\begin{tabular}{l *{8}{c}}
\toprule
\multirow{2}{*}{AOI} 
  & \multicolumn{2}{c}{MAE $\downarrow$ (m)} 
  & \multicolumn{2}{c}{RMSE $\downarrow$ (m)} 
  & \multicolumn{2}{c}{SSIM $\uparrow$} 
  & \multicolumn{2}{c}{$\text{F1}^{\text{HE}}_{\delta<1.25} \uparrow$} \\
\cmidrule(lr){2-3} \cmidrule(lr){4-5} \cmidrule(lr){6-7} \cmidrule(lr){8-9}
  & {Baseline} & {BitFit} & {Baseline} & {BitFit} & {Baseline} & {BitFit} & {Baseline} & {BitFit} \\
\midrule
Tokyo(E)     & 5.518 & \textbf{4.571} {\scriptsize (+17.2\%)} & 12.263 & \textbf{10.743} {\scriptsize (+12.4\%)} & 0.843 & \textbf{0.859} {\scriptsize (+1.9\%)} & 0.512 & \textbf{0.634} {\scriptsize (+23.9\%)} \\
Tokyo(W)     & 3.851 & \textbf{3.338} {\scriptsize (+13.3\%)} & 8.926 & \textbf{8.185} {\scriptsize (+8.3\%)} & 0.821 & \textbf{0.827} {\scriptsize (+0.7\%)} & 0.668 & \textbf{0.720} {\scriptsize (+7.8\%)} \\
Paris Core   & 6.249 & \textbf{3.829} {\scriptsize (+38.7\%)} & 8.735 & \textbf{6.059} {\scriptsize (+30.6\%)} & 0.677 & \textbf{0.774} {\scriptsize (+14.3\%)} & 0.388 & \textbf{0.758} {\scriptsize (+95.6\%)} \\
Saint-Omer   & 2.454 & \textbf{1.676} {\scriptsize (+31.7\%)} & 4.967 & \textbf{3.578} {\scriptsize (+28.0\%)} & 0.641 & \textbf{0.695} {\scriptsize (+8.4\%)} & 0.424 & \textbf{0.691} {\scriptsize (+63.0\%)} \\
S\~ao Paulo Urban    & 5.829 & \textbf{4.664} {\scriptsize (+20.0\%)} & 10.451 & \textbf{8.373} {\scriptsize (+19.9\%)} & 0.715 & \textbf{0.727} {\scriptsize (+1.6\%)} & 0.597 & \textbf{0.707} {\scriptsize (+18.4\%)} \\
S\~ao Paulo Forest   & 7.382 & \textbf{3.526} {\scriptsize (+52.2\%)} & 9.966 & \textbf{5.446} {\scriptsize (+45.4\%)} & 0.498 & \textbf{0.659} {\scriptsize (+32.4\%)} & 0.462 & \textbf{0.890} {\scriptsize (+92.6\%)} \\
\midrule
\textbf{Avg} & 5.214 & \textbf{3.600} {\scriptsize (+30.9\%)} & 9.218 & \textbf{7.064} {\scriptsize (+23.4\%)} & 0.699 & \textbf{0.757} {\scriptsize (+8.2\%)} & 0.509 & \textbf{0.734} {\scriptsize (+44.2\%)} \\
\bottomrule
\end{tabular}
}
\end{table*}
Table~\ref{tab:dav2_hrf} and Table~\ref{tab:rs3dada_bitfit} summarize the detailed results of the best-performing strategies for each backbone: HRF for Depth Anything V2 and BitFit for RS3DAda. Across six diverse AOIs, these methods consistently reduce MAE and RMSE, while substantially improving SSIM and $\text{F1}^{HE}$. The relative gains are particularly striking in challenging settings such as Saint-Omer and São Paulo (Forest), where the proposed corrections more than halve the baseline errors.

\paragraph{Qualitative validation.}
\begin{figure*}[htbp]
    \centering
    \includegraphics[width=\textwidth]{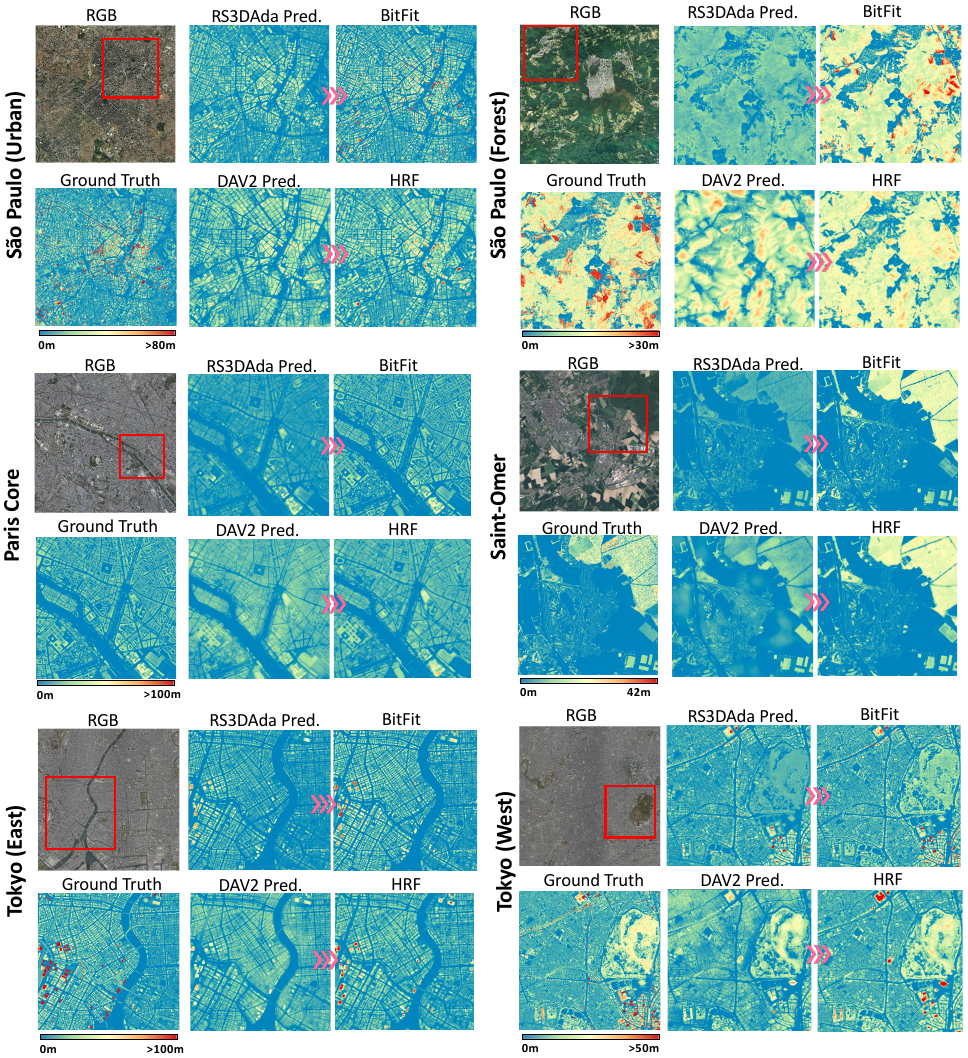}
    \caption{Qualitative results across six study areas, showing RGB images, ground truth, baseline predictions (RS3DAda and Depth Anything V2, denoted as DAV2), and their best-performing correction methods (BitFit and HRF).}
    \label{fig:qualitative}
\end{figure*}
Figure~\ref{fig:qualitative} provides qualitative comparisons across representative AOIs, juxtaposing ground truth with baseline predictions and their calibrated counterparts. The visual results reinforce the quantitative findings: HRF produces sharper building delineation on Depth Anything V2 outputs, while BitFit markedly improves RS3DAda predictions, especially in dense urban cores and forested landscapes. Together, these results demonstrate that our ICESat-2 calibration pipeline not only reduces average error but also enhances structural fidelity, yielding nDSM predictions that are both quantitatively and visually reliable.

Overall, this study establishes a flexible correction framework: HRF offers a lightweight, plug-and-play solution for general-purpose MDE models, while parameter-efficient fine-tuning (e.g., BitFit) unlocks the full potential of remote-sensing-specific MHE backbones. These insights provide actionable guidance for adapting monocular height estimation to diverse operational scenarios.

\section{Discussion and Limitations}
\label{sec:limitations_discussion}

In this work, we introduced and validated a novel, fully automated pipeline for correcting monocular height estimations using sparse ICESat-2 data. Extensive experiments over nearly 300\,km² of diverse landscapes show that our approach substantially improves the accuracy of both specialized Monocular Height Estimation (MHE) models and general-purpose Monocular Depth Estimation (MDE) models. Beyond performance, we established the first comprehensive benchmark of correction methods, revealing that the optimal strategy depends on the alignment between a model’s pre-training and the downstream task. For the domain-aligned MHE model (RS3DAda), parameter-efficient fine-tuning (PEFT) approaches such as BitFit and VPT achieved the strongest gains, while for the domain-mismatched MDE model (Depth Anything V2), an external Random Forest-based corrector proved most effective.  

Our findings also clarify the role of the pipeline. For MHE models, it is an \emph{optional yet powerful component}: when computational budgets allow and high precision is required, it can provide notable accuracy gains. For MDE models, however, the correction is \emph{indispensable}, since their outputs are inherently relative depths. Here, ICESat-2 serves as the absolute geodetic anchor needed to transform relative predictions into metrically accurate nDSMs, making the pipeline essential for practical deployment in remote sensing applications.  

A further advantage of our framework lies in its \textbf{global accessibility and scalability}. All components—including RS3DAda, Depth Anything V2, the OEM-trained semantic segmentation model, ICESat-2 photon data, and FABDEM terrain data—are open and globally available. This means that for any location worldwide, a user needs only a single georeferenced optical image to initiate the automated workflow, avoiding reliance on costly or geographically restricted commercial datasets.  

Despite these promising results and the pipeline's inherent strengths, several limitations and areas for future work should be acknowledged:

\begin{itemize}
    \item \textbf{Challenges of the Supervisory Signal:} The effectiveness of our pipeline is fundamentally tied to the ICESat-2 data. While offering global coverage, its availability, quality, and extreme sparsity pose significant challenges. Some regions may lack sufficient ground tracks, and the one-dimensional, along-track nature of the supervision signal may not be enough to resolve complex, two-dimensional error patterns, especially when fine-tuning very large models.

    \item \textbf{Computational Cost:} Although the Random Forest methods are fast, the fine-tuning approaches, even the parameter-efficient ones, require significant computational resources for training. This may limit their accessibility for users without access to high-performance GPUs, presenting a trade-off between achieving the highest possible accuracy with PEFT and the efficiency of RF-based methods.

    \item \textbf{Temporal Mismatches:} Our study used a broad time window for ICESat-2 data (2019--2024). In rapidly developing areas, a temporal gap between the optical image and the LiDAR overpass can lead to discrepancies. While our large-scale study demonstrates overall robustness, site-specific applications may require more careful temporal filtering.

    \item \textbf{Model Generalizability:} : While we demonstrated our pipeline on two state-of-the-art models (a domain-specific MHE model and a general MDE foundation model), we have not tested its applicability across the full spectrum of other available MHE/MDE architectures. Future work should validate these correction strategies on a wider variety of backbones.
\end{itemize}

In conclusion, this work demonstrates a powerful and scalable pathway for producing high-resolution 3D maps by fusing deep learning predictions with sparse satellite LiDAR. Future research should focus on integrating data from multiple sparse sources (e.g., GEDI), developing more sophisticated spatio-temporal fusion models to handle data sparsity and temporal mismatches, and also exploring self-supervised techniques to reduce the reliance on external ground-truth data altogether.

\section*{Acknowledgements}
\par This work was supported in part by the 
Japan Science and Technology Agency Fusion Oriented REsearch for disruptive Science and Technology (JST FOREST Program) 
(Grant JPMJFR206S); 
the Japan Society for the Promotion of Science (Grants 23K24865 and 24KJ0652); 
and the Next Generation AI Research Center of The University of Tokyo.

\printcredits

\bibliographystyle{cas-model2-names}

\bibliography{reference}





\end{document}